\documentclass[10pt,journal,compsoc]{IEEEtran}
\usepackage{subfigure}
\usepackage{url}
\usepackage{definitions}
\usepackage{bbm}
\usepackage{algorithm}
\usepackage{algorithmicx}
\usepackage{ragged2e}
\usepackage{caption}
\usepackage{algpseudocode}
\usepackage{booktabs}
\usepackage{graphicx}
\usepackage{wrapfig}
\usepackage{epsfig}
\usepackage{afterpage}
\usepackage{adjustbox}

\ifCLASSOPTIONcompsoc
  \usepackage[nocompress]{cite}
\else
  \usepackage{cite}
\fi

\begin{document}

\title{Hybrid Subspace Learning \\ for High-Dimensional Data}

\author{Micol~Marchetti-Bowick, %~\IEEEmembership{Student Member,~IEEE,}
        Benjamin~J.~Lengerich,\\
        Ankur~P.~Parikh, %~\IEEEmembership{Member,~IEEE,}
        and~Eric~P.~Xing%,~\IEEEmembership{Senior Member,~IEEE}% <-this % stops a space
\IEEEcompsocitemizethanks{\IEEEcompsocthanksitem M. Marchetti-Bowick and E. P. Xing are with the Machine Learning Department, Carnegie Mellon University, Pittsburgh, PA, 15213.\protect\\
% note need leading \protect in front of \\ to get a newline within \thanks as
% \\ is fragile and will error, could use \hfil\break instead.
E-mail: \{micolmb, epxing\}@cs.cmu.edu.
\IEEEcompsocthanksitem B. J. Lengerich is with the Computational Biology Department, Carnegie Mellon University, Pittsburgh, PA, 15213. \protect\\
E-mail: blengeri@cs.cmu.edu
\IEEEcompsocthanksitem A. P. Parikh is with Google Research, New York, NY, 10011.\protect\\% <-this % stops an unwanted space
E-mail: aparikh@google.com}}
%\thanks{Manuscript received December 31, 2016}}

\IEEEtitleabstractindextext{%
\begin{abstract}The high-dimensional data setting, in which p $\gg$ n, is a challenging statistical paradigm that appears in many real-world problems. In this setting, learning a compact, low-dimensional representation of the data can substantially help distinguish signal from noise. One way to achieve this goal is to perform subspace learning to estimate a small set of latent features that capture the majority of the variance in the original data. Most existing subspace learning models, such as PCA, assume that the data can be fully represented by its embedding in one or more latent subspaces. However, in this work, we argue that this assumption is not suitable for many high-dimensional datasets; often only some variables can easily be projected to a low-dimensional space. We propose a hybrid dimensionality reduction technique in which some features are mapped to a low-dimensional subspace while others remain in the original space. Our model leads to more accurate estimation of the latent space and lower reconstruction error. We present a simple optimization procedure for the resulting biconvex problem and show synthetic data results that demonstrate the advantages of our approach over existing methods. Finally, we demonstrate the effectiveness of this method for extracting meaningful features from both gene expression and video background subtraction datasets.
%demonstrate that we are able to learn an accurate low-dimensional embedding while extract extracting meaningful features.
\end{abstract}

% Note that keywords are not normally used for peerreview papers.
}

\maketitle

\IEEEdisplaynontitleabstractindextext

\IEEEpeerreviewmaketitle

\IEEEraisesectionheading{\section{Introduction}\label{sec:introduction}}

\IEEEPARstart{H}{igh-dimensional} datasets, in which the number of features $p$ is much larger than the sample size $n$, appear in a broad variety of domains. Such datasets are particularly common in computational biology \cite{marx2013biology}, where high-throughput experiments abound but collecting data from a large number of individuals is costly and impractical. In this setting, many traditional machine learning algorithms lack sufficient statistical power to distinguish signal from noise, a problem that is generally known as the curse of dimensionality~\cite{hughes1968mean}.

One way to alleviate this problem is to perform dimensionality reduction, either by choosing a subset of the original features or by learning a new set of features. In this work, we focus on the class of subspace learning methods, whose goal is to find a linear transformation that projects the high-dimensional data points onto a nearby low-dimensional subspace. This corresponds to learning a latent space representation of the data that captures the majority of information from the original features.

\begin{figure*}[ht]
\centering
 \subfigure[Data with fully low-rank structure.]{
   \includegraphics[width=0.34\linewidth]{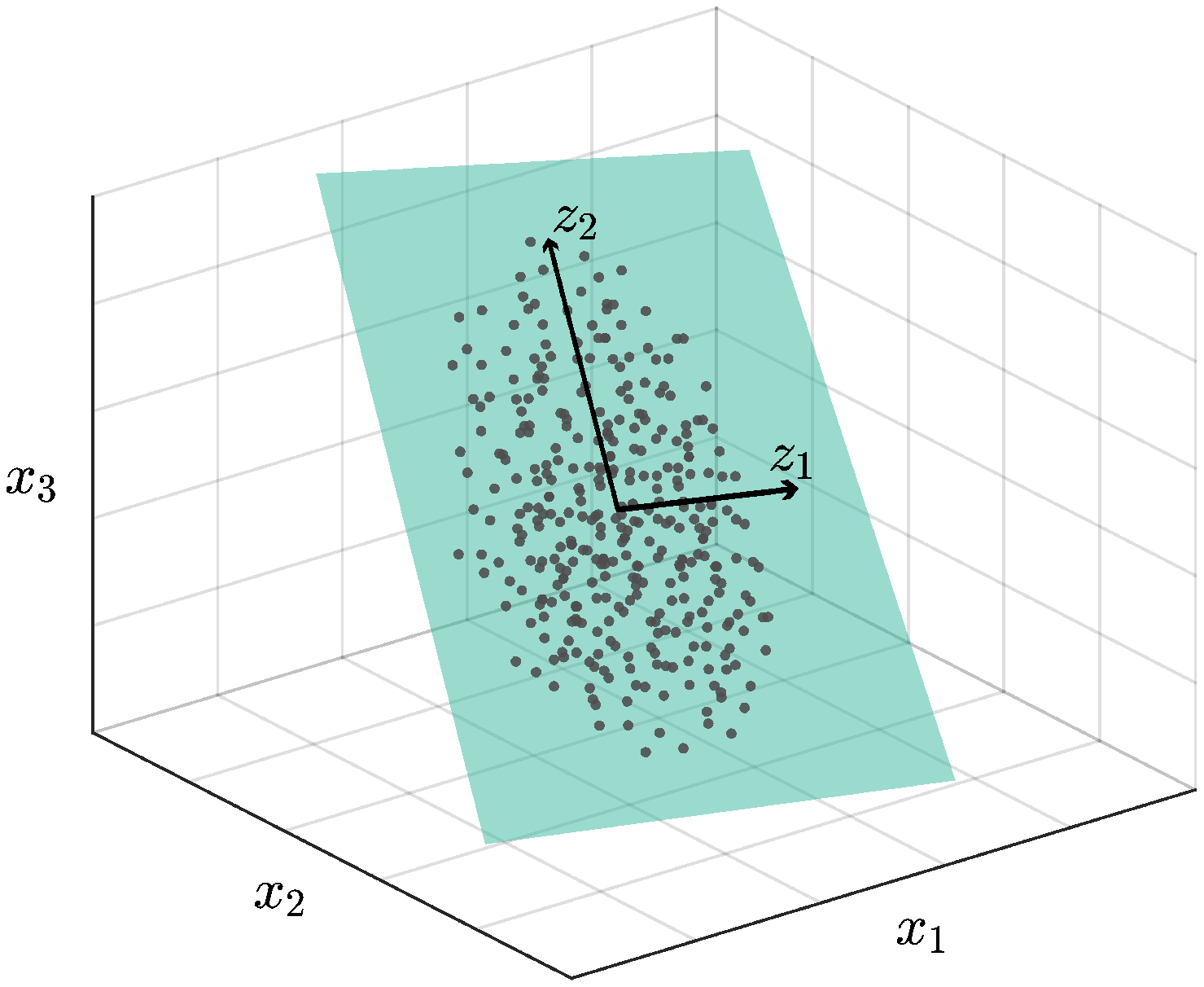}
    \label{fig:toy-data-pca}
  }  
\hspace{1cm}
\subfigure[Data with hybrid structure.]{
   \includegraphics[width=0.34\linewidth]{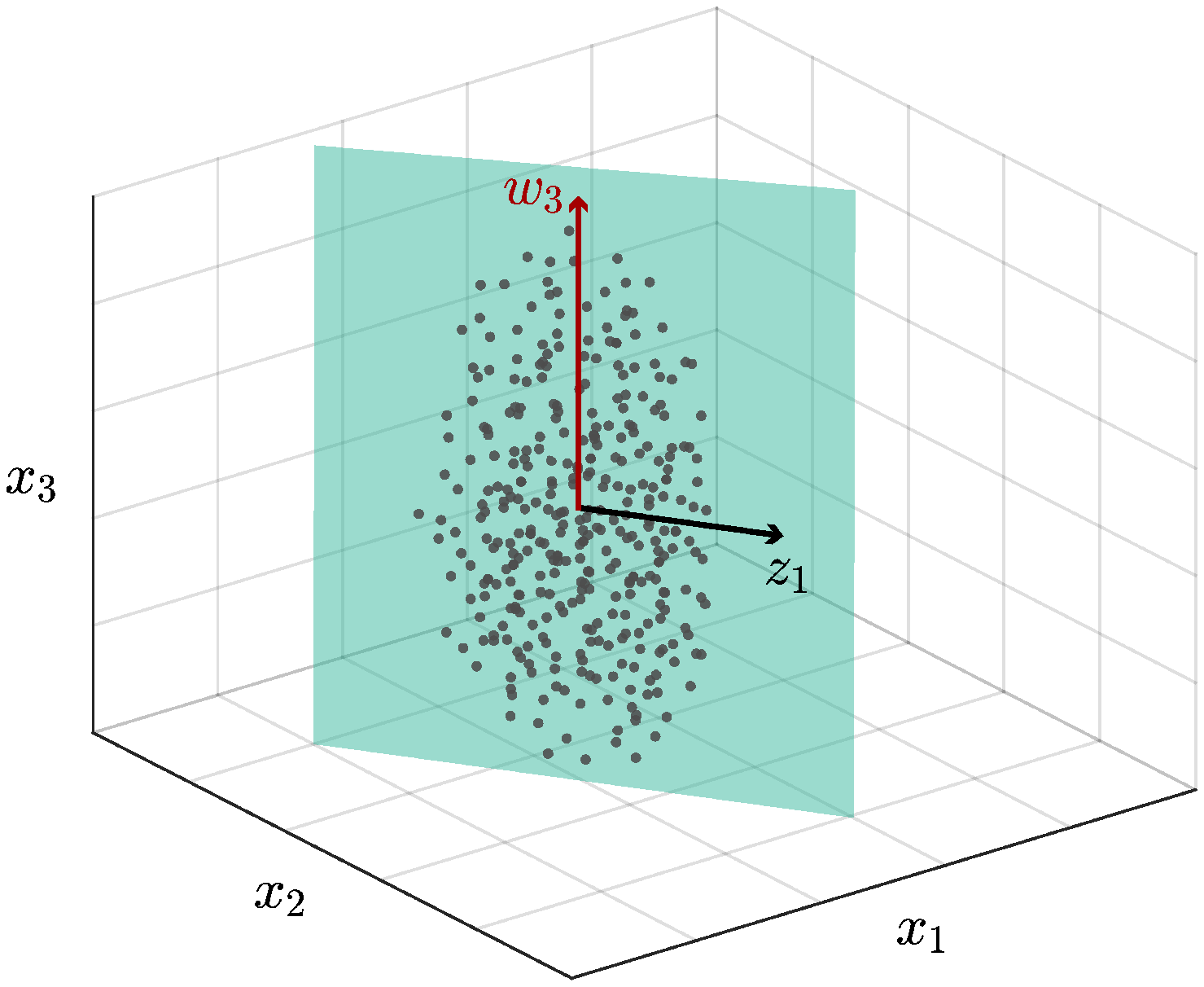}
   \label{fig:toy-data-hsl}
  }  
\caption{Toy datasets that illustrate the difference between fully low-rank data and hybrid data.}
\label{fig:toy-data}
\end{figure*}

The most popular subspace learning method is principal component analysis (PCA)~\cite{jolliffe2002principal}, which learns a compact set of linearly uncorrelated features that represent the directions of maximal variance in the original data. Since PCA was first introduced, many variants have been developed. For example, Sparse PCA~\cite{zou2006sparse} uses an elastic net penalty~\cite{zou2005regularization} to encourage element-wise sparsity in the projection matrix, resulting in more interpretable latent features. Another method, Robust PCA~\cite{candes2011robust}, learns a decomposition of the data into the sum of a low-rank component and a sparse component, which leads to increased stability in the presence of noise. Finally, there are approaches that propose richer models for the underlying latent representation of the data, involving multiple subspaces rather than just one~\cite{dorsam2007g}. %\vadjust{\newpage}

A significant limitation of nearly all existing subspace learning methods is their assumption that the data, except for noise terms, can be fully represented by an embedding in a few low-dimensional subspaces. While this may hold true in some restricted settings, we contend that in most high-dimensional, real-world datasets, only a subset of the features exhibit low-rank structure, while the remainder are best represented in the original feature space. Specifically, since the low-rank features will be highly intercorrelated, they can be accurately represented as the linear combination of a small set of latent features. However, if there are raw features that are largely uncorrelated with the others, it's clear that including them in the latent space model would require adding one new dimension for each such feature. We therefore argue that these features, which we describe as exhibiting \emph{high-dimensional} rather than \emph{low-rank} structure, should be excluded from the low-dimensional subspace representation.

We illustrate this intuition with an example. Figure~\ref{fig:toy-data} shows two toy datasets that each lie on a different 2D plane in 3D space. In the left plot, all three of the raw dimensions exhibit low-rank structure because they are all correlated. However, in the right plot, the vertical axis $x_3$ is completely uncorrelated with $x_1$ and $x_2$, which causes the 2D subspace on which the data points lie to be axis-aligned with $x_3$. We say that this data exhibits \emph{hybrid structure} because only two out of the three features are truly low-rank. 

In this simple example, PCA easily succeeds on both of the datasets shown in Figures~\ref{fig:toy-data-pca} and \ref{fig:toy-data-hsl}. However, in a high-dimensional and noisy setting, the data may not lie exactly on a low-rank subspace. In this case, we can boost the signal to noise ratio in the data by identifying a sparse set of high-dimensional features that do not contribute to the low-rank structure of the dataset and eliminating them from the low-rank projection. This is the core motivation for our approach. 

In this work, we introduce a new method called \emph{hybrid subspace learning} that estimates a latent representation of the data in which some features are mapped to a low-rank subspace but others remain in the original high-dimensional feature space. To enforce this structure, we propose a novel regularization scheme that encourages each variable to choose between participating in the low-rank or high-dimensional component of the model. The resulting problem is biconvex, and we propose an efficient alternating minimization scheme using proximal gradient descent. We further describe a warm start procedure that allows us to learn a series of increasingly penalized models while avoiding many local optima.

The goal of this method is to perform dimensionality reduction for high-dimensional datasets in a way that allows flexibility in the proportion of low-rank vs.~high-dimensional structure that is present in the data, and is also robust to noise. This allows us to learn a compact representation of the data using both \emph{feature combination} and \emph{feature selection}. As with other dimensionality reduction models, the representation that we infer can be used in various downstream tasks such as clustering and classification. Finally, we also go one step beyond dimensionality reduction by identifying a sparse set of features that stand out from the rest. Importantly, we do not assume that these features are purely noise; instead, we demonstrate that they are likely to have unique roles or functions, and therefore can provide new domain-specific insights or discoveries.
%In particular, these features are often constrained by the system under study to have low variance; their low variance is an indication of importance to the system and should not be muddied in a low-rank representation.

The remainder of this paper is organized as follows. In Section~\ref{sec:motivation}, we motivate our approach by demonstrating that certain properties of several real-world datasets naturally hint at a hybrid model. We then describe our model and optimization procedure in Sections~\ref{sec:method} and \ref{sec:opti}. We evaluate our method on synthetic data in Section~\ref{sec:synth-exp} and on two real-world datasets in Section~\ref{sec:real-exp}. Finally we conclude in Section~\ref{sec:conclusion} by discussing the implications of our findings. 

\textbf{Notation}: We use lowercase bold symbols for vectors $\xb$ and uppercase bold symbols for matrices $\Xb$. The $i^\text{th}$ element of $\xb$ is denoted $\xb(i)$, the $i^\text{th}$ row and $j^\text{th}$ column of $\Xb$ are denoted $\Xb(i,:)$ and $\Xb(:,j)$, respectively, and $\diag(\xb)$ denotes a diagonal matrix $\Xb$ s.t.~$\Xb(i,i) = \xb(i)$. We use $\| \cdot \|_1$ for the element-wise $l_1$ norm of a vector or matrix, $\| \cdot \|_2$ for the $l_2$ norm of a vector, $\| \cdot \|_F$ for the Frobenius norm of a matrix, and $\| \cdot \|_{1,p}$ to denote an $l_{1,p}$ column-wise block norm of a matrix s.t.~$\| \Xb \|_{1,p} = \sum_{j} \| \Ab(:,j) \|_p$. 

\section{Motivation}
\label{sec:motivation}

In this section, we demonstrate via a series of simulations that hybrid structure causes the singular value spectrum of a dataset to become ``long-tailed" i.e.~to have a distribution in which a large amount of the probability mass is far from the mean. We then show that many real-world biological datasets possess long-tailed singular value spectra, which implies that it is not appropriate to attempt to capture all of the variables with a low-dimensional representation.

\begin{figure*}[tbp]
 \centering
 \subfigure[real data 1]{
     \includegraphics[width=0.16\linewidth]{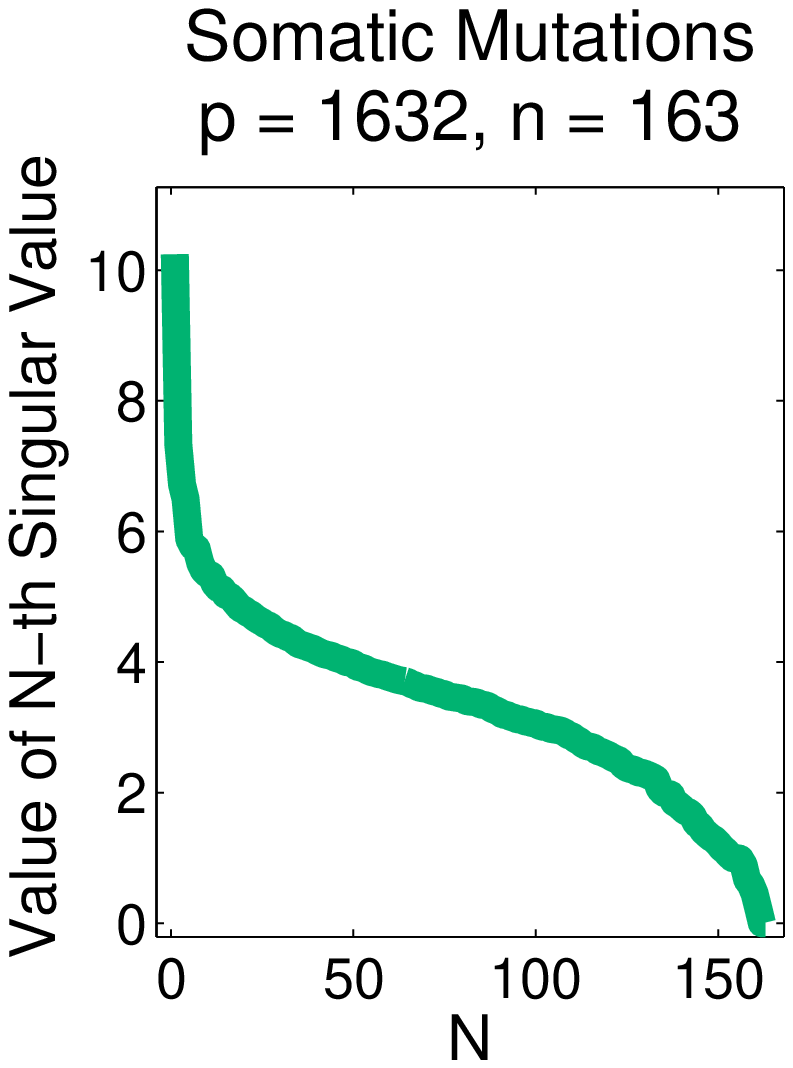}
      \label{fig:real-data-ssm}
  }  
   \subfigure[real data 2]{
     \includegraphics[width=0.16\linewidth]{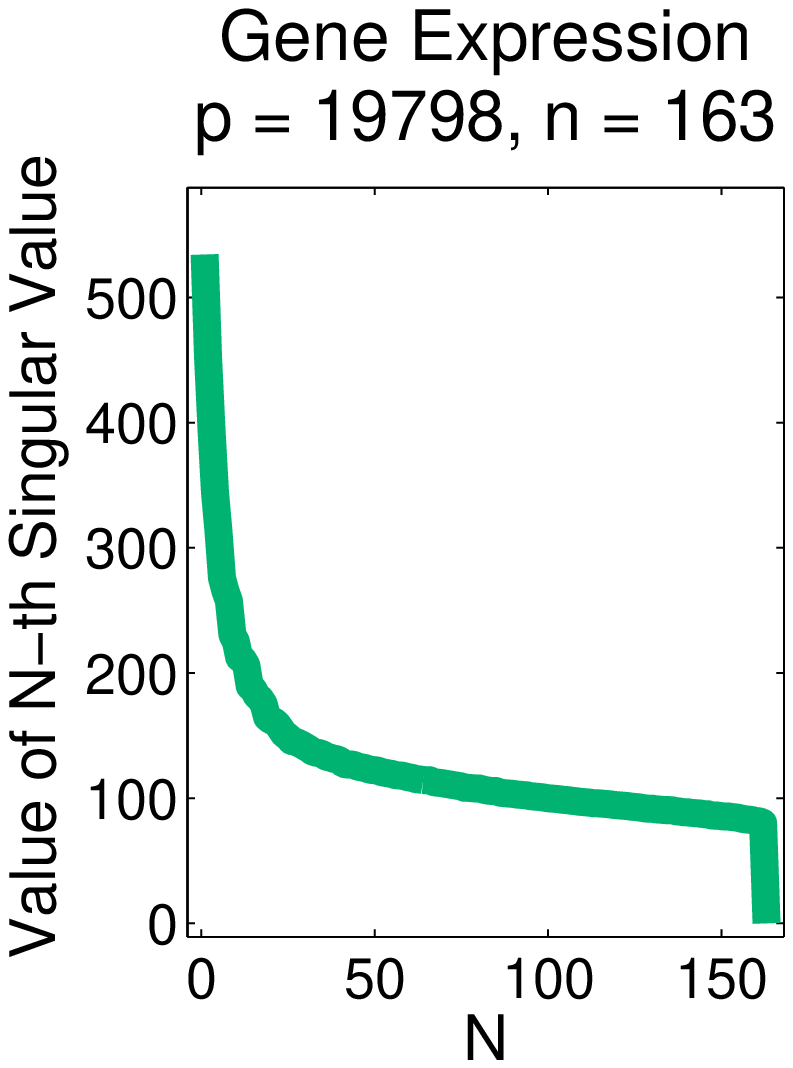}
      \label{fig:real-data-gexp}
  }  
   \subfigure[real data 3]{
     \includegraphics[width=0.16\linewidth]{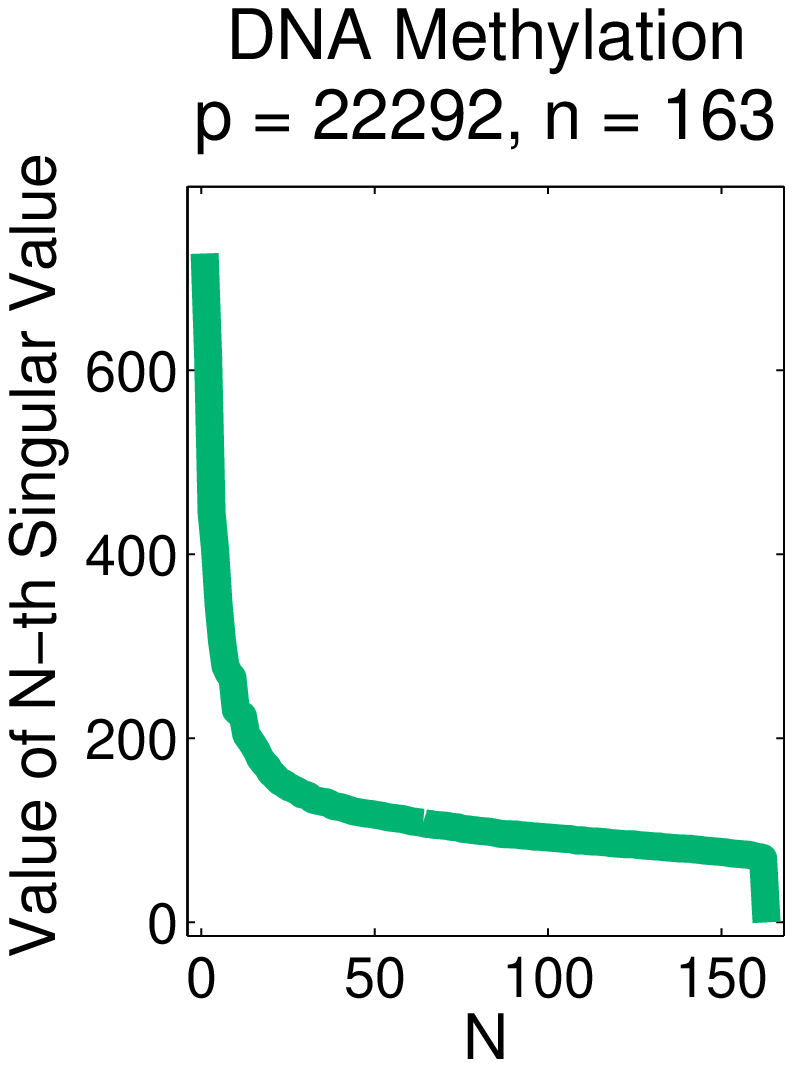}
      \label{fig:real-data-dmeth}
  }  
     \subfigure[real data 4]{
     \includegraphics[width=0.16\linewidth]{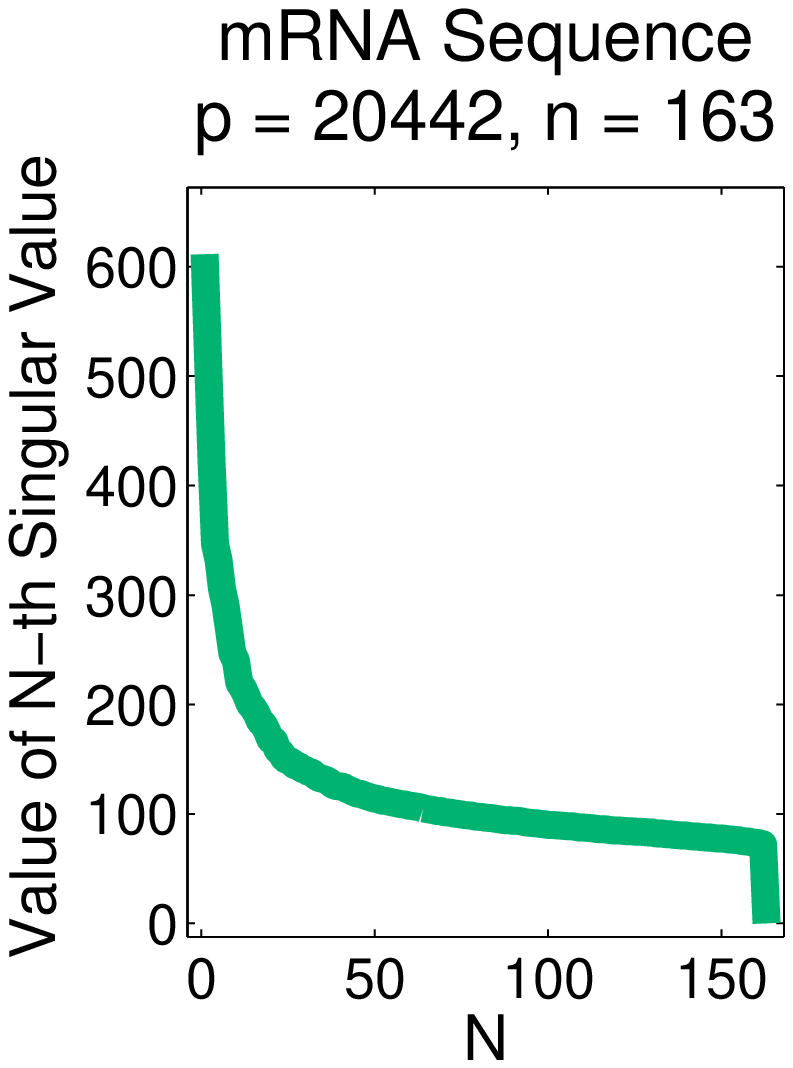}
      \label{fig:real-data-rna}
  }  
     \subfigure[real data 5]{
     \includegraphics[width=0.16\linewidth]{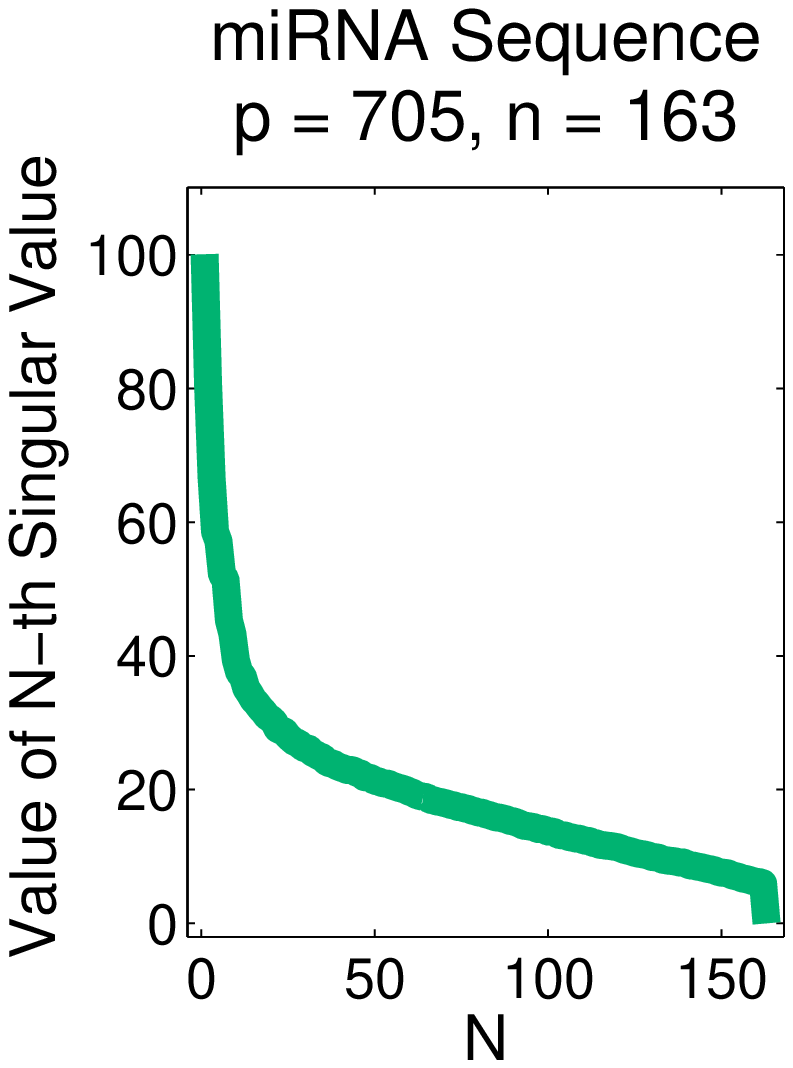}
      \label{fig:real-data-mirna}
  }  
  
   \subfigure[$\theta=(1,0,0)$]{
     \includegraphics[width=0.16\linewidth]{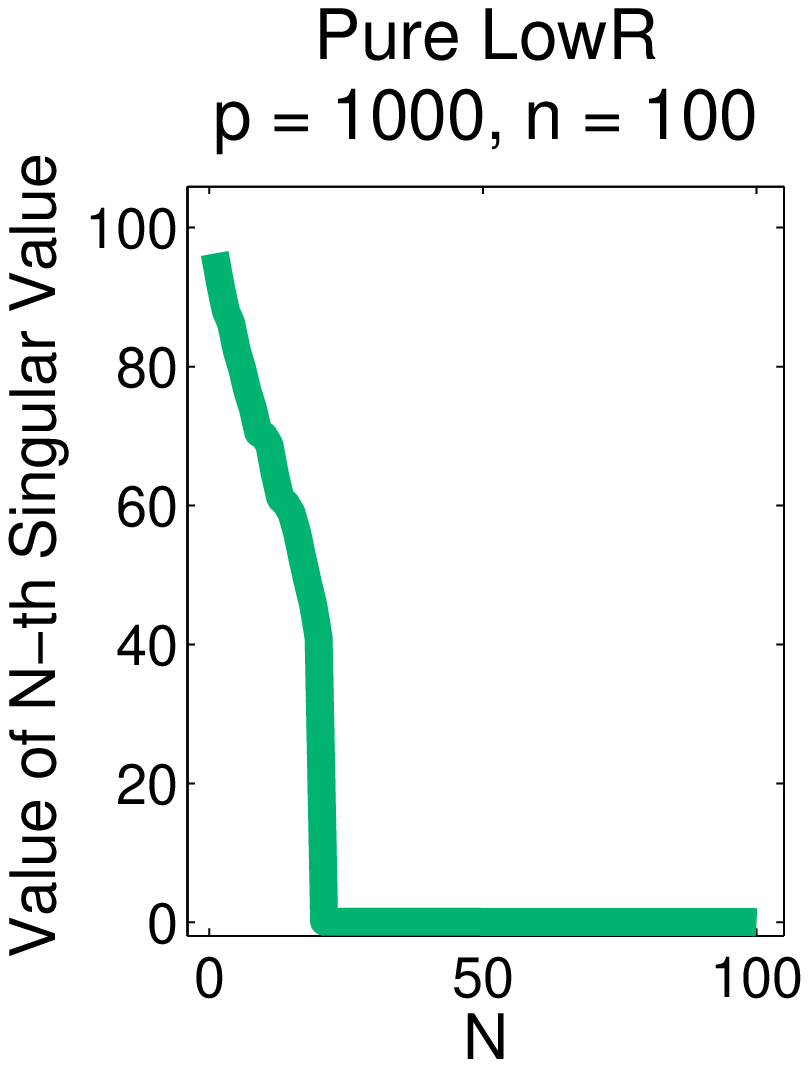}
      \label{fig:synth-data-1-0-0}
  }  
     \subfigure[$\theta=(0,1,0)$]{
     \includegraphics[width=0.16\linewidth]{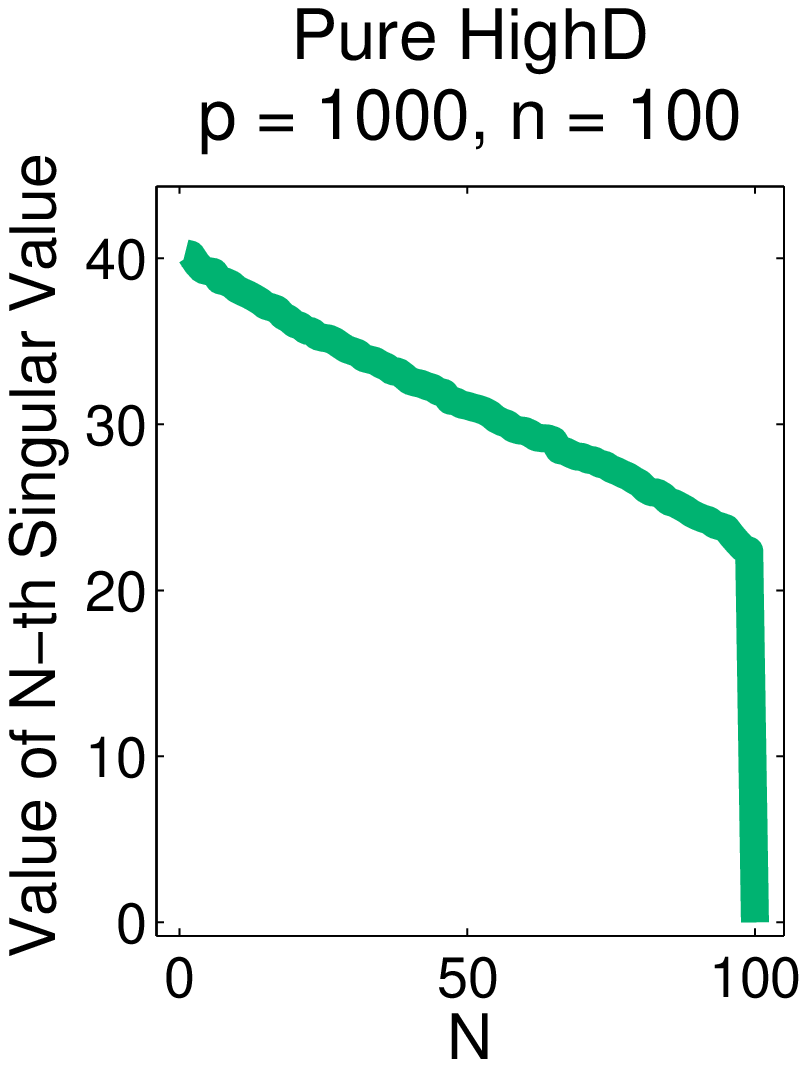}
      \label{fig:synth-data-0-1-0}
  }  
   \subfigure[$\theta=(.8,.2,0)$]{
     \includegraphics[width=0.16\linewidth]{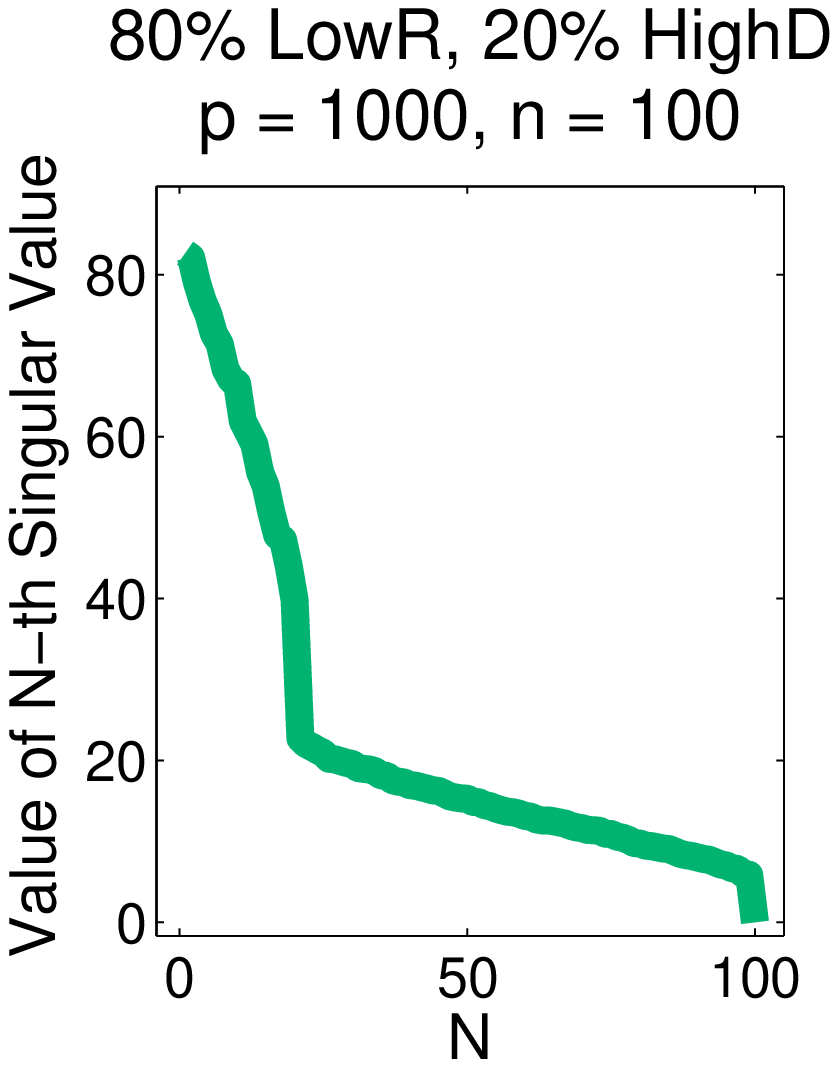}
      \label{fig:synth-data-8-2-0}
  }  
   \subfigure[$\theta=(.5,.5,0)$]{
     \includegraphics[width=0.16\linewidth]{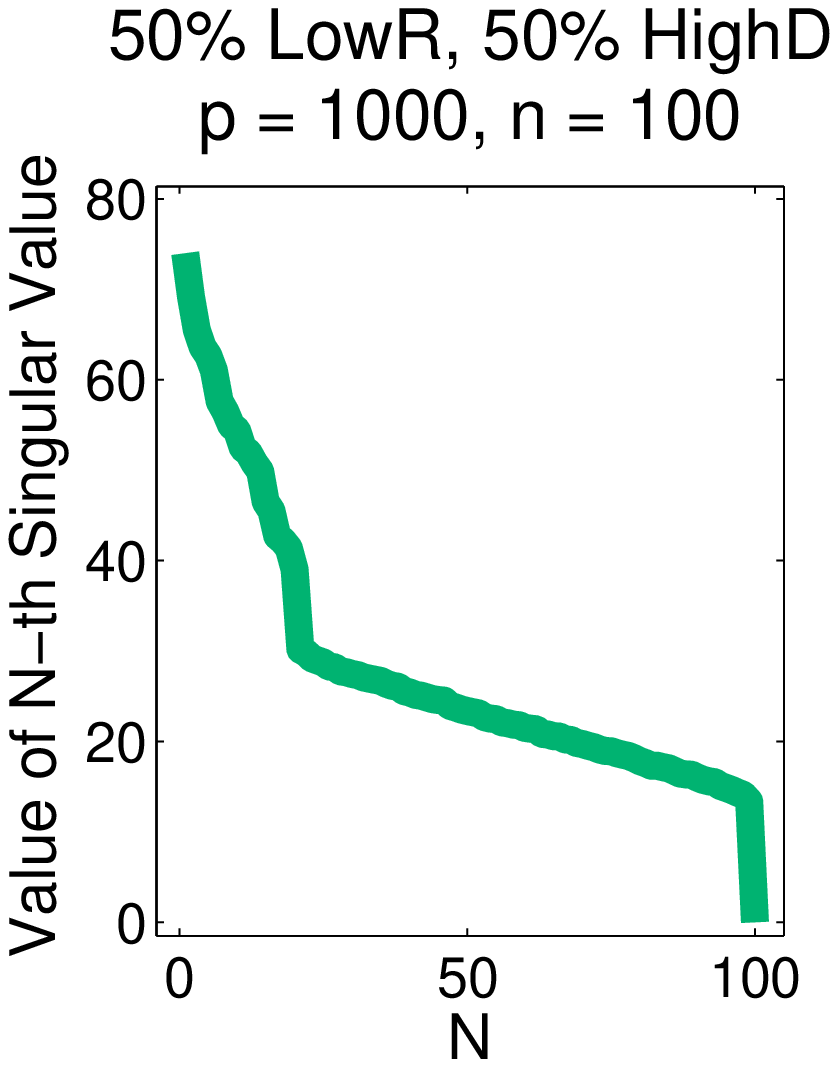}
      \label{fig:synth-data-5-5-0}
  }  
     \subfigure[$\theta=(.2,.8,0)$]{
     \includegraphics[width=0.16\linewidth]{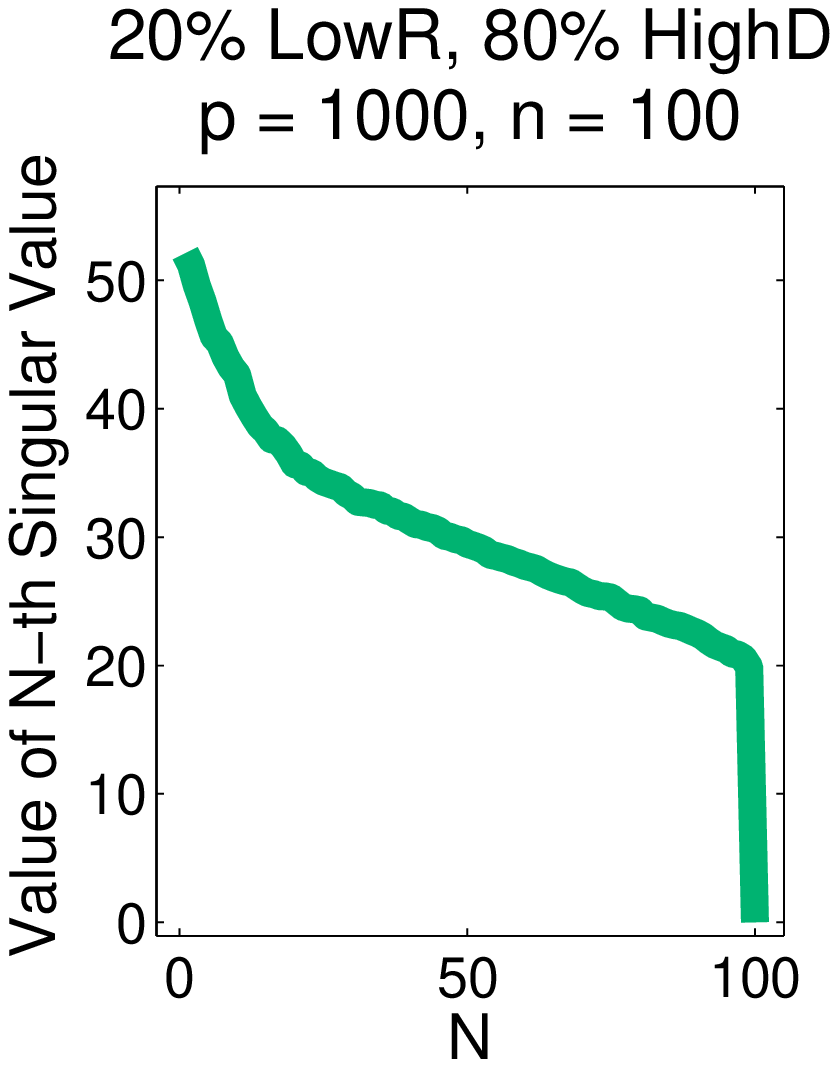}
      \label{fig:synth-data-2-8-0}
  }  
 \caption[]{Singular value spectra of real and synthetic datasets; (a)-(e) five real biological datasets collected from tumor samples of 163 leukemia patients; (f) synthetic data with pure low-rank structure; (g) synthetic data with pure high-dimensional structure; (h)-(j) synthetic data with hybrid structure.} %\shaycomment{fix trees}
  \label{fig:spectrum-plots}
\end{figure*}

Consider a dataset $\Xb \in \mathbb{R}^{n \times p}$ with $n$ samples and $p$ features. 
The top row of Figure~\ref{fig:spectrum-plots} shows the singular value spectra of five real datasets that consist of measurements taken from tumor samples of cancer patients. In all of these datasets, the top singular values are large but then decay very quickly. However, instead of going directly to zero, the spectrum has a long tail. This indicates the presence of structure in the data that does not fit into a low-rank space. As a result, if we ignored the tail by projecting the data to a low-rank subspace, it is likely that we would only capture a very coarse-grained representation of the data.

We compare these real datasets with several simulated datasets to demonstrate how certain underlying modeling assumptions affect the singular value spectrum of the data. We generate synthetic data as follows. Let $\Zb$ be an $n \times k$ matrix with full column rank, $\Ab$ be a $k \times p$ matrix with full row rank, and $\Wb$ be an $n \times p$ matrix whose elements are independent. Define a probability vector $\theta = (\theta_1, \theta_2, \theta_3)$ that specifies the likelihood that each feature participates in only a low-rank (low-r) component, only a high-dimensional (high-d) component, or both, respectively. For simplicity, we consider only the case of $\theta_3 = 0$ for now. For each variable $j \in \{1,\ldots,p\}$, we draw $C_j \sim \textrm{Categorical}(\theta)$. If $C_j = (1,0,0)$, we set $\Xb(:,j) \sim \mathcal{N}(\Zb \Ab(:,j), \sigma^2 \Ib_{n\times n})$. If $C_j = (0,1,0)$, we set $\Xb(:,j) \sim \mathcal{N}(\Wb(:,j), \sigma^2 \Ib_{n\times n})$. 

For our simulations, we use $n=100$, $p=1000$, $k=20$, and $\sigma^2$ close to $0$, and plot the spectra of five synthetic datasets generated for multiple values of $\theta$ in the bottom row of Figure~\ref{fig:spectrum-plots}. In panel (f), we set $\theta = (1,0,0)$ such that $\Xb$ is rank $k$ with some random noise. In this case the singular value spectrum drops sharply after $k$, but the tail that appears in the real data is missing. While it is possible that the tail could only contain noise, we postulate that it contains some important information that is ignored by subspace learning methods that focus purely on low-rank structure. In panel (g), we set $\theta = (0,1,0)$ such that $\Xb$ has rank $n$. In this case, the singular value spectrum of $\Xb$ decays slowly, again unlike the real data. This implies that methods that use the full data matrix $\Xb$ without alteration are not exploiting its intrinsic structure. 

Panels (h)-(j) display three ``hybrid'' settings of $\theta$. The spectra of these datasets exhibit a structure that is much more similar to the real data, with a few large singular values, and a tail that decays slowly. In these cases, This is the motivation for our hybrid approach that can model both the head and tail of the singular value spectrum.%\footnote{We note that in cases (h)-(j), attempting to force all of the variables to fit into a low-dimensional space would necessitate including a large number of dimensions in the subspace, many of which would be highly underutilized.}

%\vspace{-.4cm}

%\begin{methods}

\section{Model}
\label{sec:method}

%In this section, we propose our novel approach for hybrid multi-view subspace learning. 
Given a dataset $\Xb \in \mathbb{R}^{n \times p}$, traditional subspace learning aims to solve the following problem: 
\begin{flalign}
 \min_{\Zb, \Ab} 
 ~ \textstyle \| \Xb -  \Zb \Ab \|_F^2
 \label{eqn:vanilla}
 \end{flalign}
where $\Zb \in \RR^{n \times k }$ is a $k$-dimensional representation of each point and $\Ab \in \RR^{k \times p}$ is a transformation that maps the latent space to the observed feature space.  The above model, which is equivalent to PCA when the columns of $\Zb$ are constrained to be orthogonal, implicitly assumes that all of the information in $\Xb$ can be captured by its embedding in a low-rank subspace. %However, there have been a series of extensions that have focused on a similar but more relaxed assumption that $\Xb$ can be fully represented by a combination of shared and view-specific low-rank latent spaces~\cite{jia2010factorized,archambeau2008sparse,virtanen2011bayesian,virtanen2011factor}. 
%A significant drawback of this method, along with other existing matrix factorization techniques, such as Robust PCA and Sparse PCA, is that they attempt to force all of the features in $\Xb$ to fit into a low-rank representation. 
However, as previously discussed, this assumption is inappropriate for high-dimensional data with a long-tailed singular value spectrum.

%We seek to augment this loss function to reflect our intuition in the previous section.  %the view-specific information is instead .  In addition, each view has a high-d component that is sparsely used to be able to represent the long tail of  the singular value spectrum. We assume that most of the variables exist only in the low-rank or high-d components, but a few are allowed to participate in both. This leads naturally to the following loss function:
To overcome this limitation, we propose a new, flexible model for subspace learning that allows each feature in $\Xb$ to choose between participating in a low-rank representation, $\Zb$, or a high-dimensional representation, $\Wb$. 
%Here our assumption is that the information common to all views can be captured by a low-rank structure, but a subset of the features in each view will not fit in this structure and can be better modeled by a high-d component. 
With this formulation, the goal is to have the low-r component capture the head of the singular value spectrum while the high-d component captures the tail. This leads naturally to the following problem:
\begin{align}
\min_{\Zb, \Ab, \Wb, \bb}   
~& \textstyle \| \Xb -  \Zb \Ab - \Wb \diag(\bb) \|_F^2 +  \lambda \textstyle \|\bb\|_0 \notag \\%[5pt]
\textrm{s.t.   } ~~~ &\|\Ab(:,j)\|_2 \cdot \bb(j) = 0 ~~~ \forall \, j \notag \\ & \| \Wb \|_F \leq 1 
\label{eqn:hard-obj} 
\end{align}
%\begin{align}
%\min_{\Zb, \Ab, \Wb, \bb}   
%~& \textstyle \| \Xb -  \Zb \Ab - \Wb \|_F^2 +  \lambda \textstyle \|\Wb\|_{1,0} \notag \\%[5pt]
%\textrm{s.t.   } ~~~ &\|\Ab(:,j)\|_2 \cdot \| \Wb(:,j) \|_{2} = 0 ~~~ \forall \, j
%\label{eqn:hard-obj} 
%\end{align}
Here, $\Zb \in \RR^{n \times k}$ is the low-rank component (as before) and $\Wb \in \RR^{n \times p}$ is the high-dimensional component. Furthermore, $\bb \in \{0,1\}^{p}$ is a vector of indicator variables, each of which dictates whether or not a particular feature $j$ participates in the high-d component. We apply an $\ell_0$ norm regularizer to restrict the total number of features that are captured by the high-d component. % Without this penalty, the model would be useless because every feature could be represented exactly by $\Wb$.
Finally, we constrain the problem such that each feature belongs to exactly one component. 

However, this problem is intractable for two reasons. First, the $\ell_0$ penalty is highly nonconvex and difficult to optimize. Secondly, since $\Ab$ and $\bb$ are coupled in the constraint, they cannot be optimized jointly. Performing alternating minimization on (\ref{eqn:hard-obj}) would yield degenerate solutions, since initializing $\bb(j)$ to non-zero would always force $\Ab(:,j)$ to be zero and vice-versa. We therefore propose the following relaxation:
\begin{flalign}
\min_{\Zb, \Ab, \Wb, \bb}  
~ & \textstyle \| \Xb -  \Zb \Ab - \Wb \diag(\bb) \|_F^2  \notag \\[-5pt]
& + \gamma \| \Ab \diag(\bb)\|_{1,2} + \lambda \textstyle \|\bb\|_1 \notag \\[5pt]
\textrm{s.t.   } ~~~ &\| \Zb \|_F \leq 1 ~~~  \| \Wb \|_F \leq 1 
\label{eqn:hybrid-obj} 
\end{flalign}
We make two changes in order to arrive at (\ref{eqn:hybrid-obj}). First, as is common in the sparsity literature, we relax $\bb \in \{0,1\}^{p}$ to $\bb \in \RR^{p}$, and replace the $l_0$ penalty on $\bb$ with an $l_1$ penalty. Second, and more unique to our problem, we replace the hard constraint on $\Ab$ and $\bb$ in (\ref{eqn:hard-obj}) with a structured sparse regularizer that encourages each feature to participate in either the low-r component ($\Zb$) or the high-d component ($\Wb$), but not both. This is achieved with an $l_{1,2}$ norm penalty over $\Ab$ and $\bb$ of the form $\| \Ab \diag(\bb) \|_{1,2} = \sum_{j=1}^{p} \bb(j) \|\Ab(:,j)\|_2$. Notice that sparsifying either the $j$th element of $\bb$ or the $j$th column of $\Ab$ will completely zero out the $j$th term of the penalty. This regularization scheme therefore encourages mutually exclusive sparsity over the columns of $\Ab$ and the elements of $\bb$. Furthermore, once the $j$th term of the penalty is zero, there is no longer any shrinkage applied to the $j$th column of $\Ab$, which yields a better estimate of the model parameters and eliminates the need for refitting the low-rank model after the high-d features have been identified.

As $\gamma$ tends to $\infty$, the model shown in (\ref{eqn:hybrid-obj}) will enforce the hard constraint in (\ref{eqn:hard-obj}). Conveniently, as we will see in the next section, this relaxation also permits us to develop a much more effective optimization procedure that is less likely to be trapped in local optima. At the same time, the new model is more flexible than (\ref{eqn:hard-obj}) in that it can allow some overlap between $\Ab$ and $\bb$ at the cost of having an additional tuning parameter. %  In particular, as described in the next section, we set $\gamma$ to zero and then gradually increase it, using the warm starts approach of~\cite{mazumder2011sparsenet} to initialize each successive model.

%Note that some approaches~\cite{} may add additional view specific low-rank latent spaces~\cite{}, which we omit for simplicitly, but can also be easily incorporated into our approach while preserving the biconvexity.

Our approach, hybrid subspace learning (HSL), is closely related to Robust PCA (RPCA)~\cite{candes2011robust} and its variants, which learn a decomposition of the data $\Xb$ into the sum of a low-rank component $\Lb$ and a sparse component $\Sb$. In particular, while RPCA encourages element-wise sparsity in $\Sb$, Outlier Pursuit (OP)~\cite{xu2010robust} is a more structured approach that encourages row-wise sparsity in $\Sb$ in order to identify points in the dataset that are outliers, and allow them to be ignored by the low-rank representation $\Lb$. The OP algorithm can just as easily be applied to a transposed data matrix to identify features that are ``outliers'' because they can't easily be embedded in a low-rank subspace. %Similarly, HSL can be seen as identifying features that are outliers and pushing them into the high-dimensional component. 
Although this is conceptually very similar to the core idea of HSL, there are several key differences. 

First, and most importantly, HSL also strictly enforces sparsity in the projection matrix $\Ab$, which causes some features to be completely excluded from the low-rank representation. In OP, although $\Sb$ can be made column-wise sparse, there is nothing to prevent the features that participate in $\Sb$ from also participating in $\Lb$. Second, we learn an exact rank $k$ low-rank representation, whereas OP aims to minimize the nuclear norm of $\Lb$. Finally, HSL also has connections to Sparse PCA (SPCA)~\cite{zou2006sparse}, which learns a rank $k$ decomposition of $\Xb$ given by $\Zb \Ab$, where $\Ab$ is encouraged to be element-wise sparse. %We compare our method to each of the approaches discussed above in Section~\ref{sec:synth-exp}.

%\vspace{-.4cm}

%\begin{methods}
\section{Optimization}
\label{sec:opti}

Our optimization objective consists of a differentiable, biconvex loss function, $$\ell(\Zb, \Ab, \Wb, \bb) = \| \Xb -  \Zb \Ab - \Wb \diag(\bb) \|_F^2$$ and two non-smooth, biconvex regularizers, $$\psi(\Ab, \bb) = \|\Ab \diag(\bb) \|_{1,2}  ~~ \text{ and } ~~ \phi(\bb) = \|\bb\|_1.$$ The objective is jointly convex in $\{\Wb, \Ab\}$ when $\Zb$ and $\bb$ are fixed, and is jointly convex in \{$\Zb, \bb$\} when $\Wb$ and $\Ab$ are fixed. We implement an alternating minimization scheme to solve this problem, in which we iteratively optimize each convex sub-problem until the complete objective converges. Since the objective function of each sub-problem consists of a smooth, convex loss function plus a non-smooth, convex regularizer, we can leverage well-known tools to optimize functions of this form. Specifically, we apply proximal gradient descent, which projects the gradient step back onto the solution space at each iteration. The complete optimization procedure is outlined in Algorithm~\ref{algo:opt-proxgrad}. In practice, we employ accelerated proximal gradient descent with line search to achieve a convergence rate of $O({1}/{\sqrt{\epsilon}})$~\cite{beck2009fast}. We also find that 25-50 outer iterations is typically sufficient to reach convergence.  

The projection and proximal operators used on lines 8, 10, 16, and 18 of Algorithm \ref{algo:opt-proxgrad} are defined as:
\begin{flalign}
& l_F\text{--project}\big( \Wb \big) = \Wb / \max \{ 1, \| \Wb \|_F \} \label{eqn:l2project}  \\[-1pt]
%& l_2\text{--project}\big( \wb \big) = \wb / \max \{ 1, \| \wb \|_2 \} \label{eqn:l2project}  \\[-1pt]
& l_2\text{--prox} \big( \ab , u \big) = \ab \cdot \max \{ 0, \|  \ab \|_2 - u \} / \| \ab \|_2  \label{eqn:l2prox}  \\[-1pt]
& l_1\text{--prox} \big( b , u \big) = \sgn(b) \cdot \max \{ 0, | b | - u \} \label{eqn:l1prox}
\end{flalign}
These are applied column-wise or element-wise when given matrix arguments in place of vectors or vector arguments in place of scalars, respectively. We also use $|\bb|$ to denote the element-wise absolute value of $\bb$, and $\|\Ab\|_{\cdot,2}$ to denote the column-wise $l_2$ norm of $\Ab$.

% We can then rewrite the optimization problem given in (\ref{eqn:hybrid-obj}) as follows:
%\begin{flalign}
%\min_{{\Zb, \Ab, \Wb, \bb}}  ~~ & \| \Xb -  \Zb \Ab - \Wb \diag(\bb) \|_F^2  + \lambda \|\bb\|_1 + \gamma \| \Ab \diag(\bb)\|_{1,2} \notag \\
%\textrm{s.t.   } ~~~~~~~ &\| \Zb(:,j) \|_2 \leq 1 ~~~ \forall \, j  \hspace{.5cm} \| \Wb(:,j) \|_2 \leq 1 ~~~ \forall \, j 
%\label{eqn:simple-hybrid-obj}
%\end{flalign}

\begin{algorithm}[tbp]
{\fontsize{9}{13}\selectfont
\caption{Proximal Gradient Descent for HSL}
\begin{algorithmic}[1]
\State \textbf{inputs:} data matrix $\Xb$; regularization parameters $\lambda$, $\gamma$; step size $\alpha$; initial values $\Zb$, $\Ab$, $\Wb$, $\bb$
\State initialize $\hat{\Zb}$, $\hat{\Ab}$, $\hat{\Wb}$, $\hat{\bb}$ using provided initial values
\Repeat 
\State fix \smash{$\Zb = \hat{\Zb}$, $\bb = \hat{\bb}$}
\State initialize \smash{$\Wb^0 = \hat{\Wb}$, $\Ab^0 =\hat{\Ab}$}
\Repeat \Comment{Optimize $\{\Wb,\Ab\}$ }
\State \smash{$\Wb^+ = \Wb^t - \alpha \, \nabla_{\Wb} \, \ell (\Zb,\Ab^t,\Wb^t,\bb)$}
\State \smash{$\Wb^{t+1} = \text{$l_F$--project} \big( \Wb^+ \big)$} \Comment{Eq. (\ref{eqn:l2project})}
\State \smash{$\Ab^+ = \Ab^t - \alpha \, \nabla_{\Ab} \, \ell (\Zb,\Ab^t,\Wb^t,\bb)$}
%\State \smash{$\ub = \alpha \left( \lambda + \gamma |\bb| \right)$}
\State \smash{$\Ab^{t+1} = \text{$l_2$--prox} \big( \Ab^+,  \alpha \, \gamma |\bb| \big)$} \Comment{Eq. (\ref{eqn:l2prox})}
\Until{convergence}
\State fix \smash{$\Wb = \hat{\Wb}$, $\Ab = \hat{\Ab}$}
\State initialize \smash{$\Zb^0 = \hat{\Zb}$, $\bb^0 = \hat{\bb}$}
\Repeat \Comment{Optimize $\{\Zb,\bb\}$}
\State \smash{$\Zb^+ = \Zb^t  - \alpha \, \nabla_{\Zb} \, \ell(\Zb^t,\Ab,\Wb,\bb^t)$}
\State \smash{$\Zb^{t+1} = \text{$l_F$--project} \big( \Zb^+ \big)$} \Comment{Eq. (\ref{eqn:l2project})}
\State \smash{$\bb^+ =  \bb^t - \alpha \, \nabla_{\bb} \, \ell(\Zb^t,\Ab,\Wb,\bb^t)$}
%\State \smash{$\ub = \alpha \left( \gamma \| \Ab \|_{\cdot,2} \right)$}
\State \smash{$\bb^{t+1} = \text{$l_1$--prox} \big( \bb^+, \alpha \left( \gamma \| \Ab \|_{\cdot,2}  + \lambda \right) \big)$} \Comment{Eq. (\ref{eqn:l1prox})}
\Until{convergence}
\Until{convergence}
\State \textbf{outputs:} estimates $\hat{\Zb}$, $\hat{\Ab}$, $\hat{\Wb}$, $\hat{\bb}$
\end{algorithmic}
\label{algo:opt-proxgrad}
}
\end{algorithm}

\begin{algorithm}[tbp]
{\fontsize{9}{13}\selectfont
\caption{Warm Starts for HSL}
\begin{algorithmic}[1]
\State \textbf{inputs:} data matrix $\Xb$; regularization parameter $\lambda$; increment size $\eta$; step size $\alpha$
\State randomly initialize $\Zb^0$, $\Ab^0$, $\Wb^0$, $\bb^0$
\State initialize $\gamma = 0$
\While{$\| \Ab \diag(\bb)\|_{1,2} > 0$}
\State \smash{\big($\Zb^{i+1}$, $\Ab^{i+1}$, $\Wb^{i+1}$, $\bb^{i+1}$)}
\item[] \quad ~~~~~ $\leftarrow$ \smash{ ProxGD-HSL \big($\Xb$, $\lambda$, $\gamma$, $\alpha$, $\Zb^{i}$, $\Ab^{i}$, $\Wb^{i}$, $\bb^{i}$ \big)}
\State update $\gamma \leftarrow \gamma + \eta$
\EndWhile
\State \textbf{outputs:} final estimates of $\Zb$, $\Ab$, $\Wb$, $\bb$ % $\hat{\Zb}$, $\hat{\Ab}$, $\hat{\Wb}$, $\hat{\bb}$
\end{algorithmic}
\label{algo:opt-warmstart}
}
\end{algorithm}

Although this optimization procedure is quite efficient, the algorithm can easily get trapped in local optima. The joint regularization term compounds the problem by increasing the sensitivity of the algorithm to initialization, especially when the value of $\gamma$ is very high. However, when $\gamma$ is small, these local optima are substantially reduced. Therefore, to circumvent this problem, we fit our model to data by incrementally increasing the value of $\gamma$ from 0 to $\gamma_{\text{max}}$, while using warm starts to initialize the estimate of each successive model.\footnote{This is based on ideas by~\cite{mazumder2011sparsenet} who proposed warm starts for a non-convex sparse regularizer.} In the next section, we demonstrate empirically that using warm starts in place of cold starts leads to significant performance gains. The warm starts procedure is shown in Algorithm~\ref{algo:opt-warmstart}, where we define $\gamma_{\text{max}}$ as the smallest value of $\gamma$ that yields $\| \Ab \diag(\bb)\|_{1,2} = 0$. %Furthermore, the joint regularization term compounds the problem by increasing the sensitivity of the algorithm to initialization, especially when the value of $\gamma$ is very high.

\section{Synthetic Data Experiments}
\label{sec:synth-exp}

\begin{figure*}[tbp]
\centering
   \includegraphics[width=0.8\linewidth]{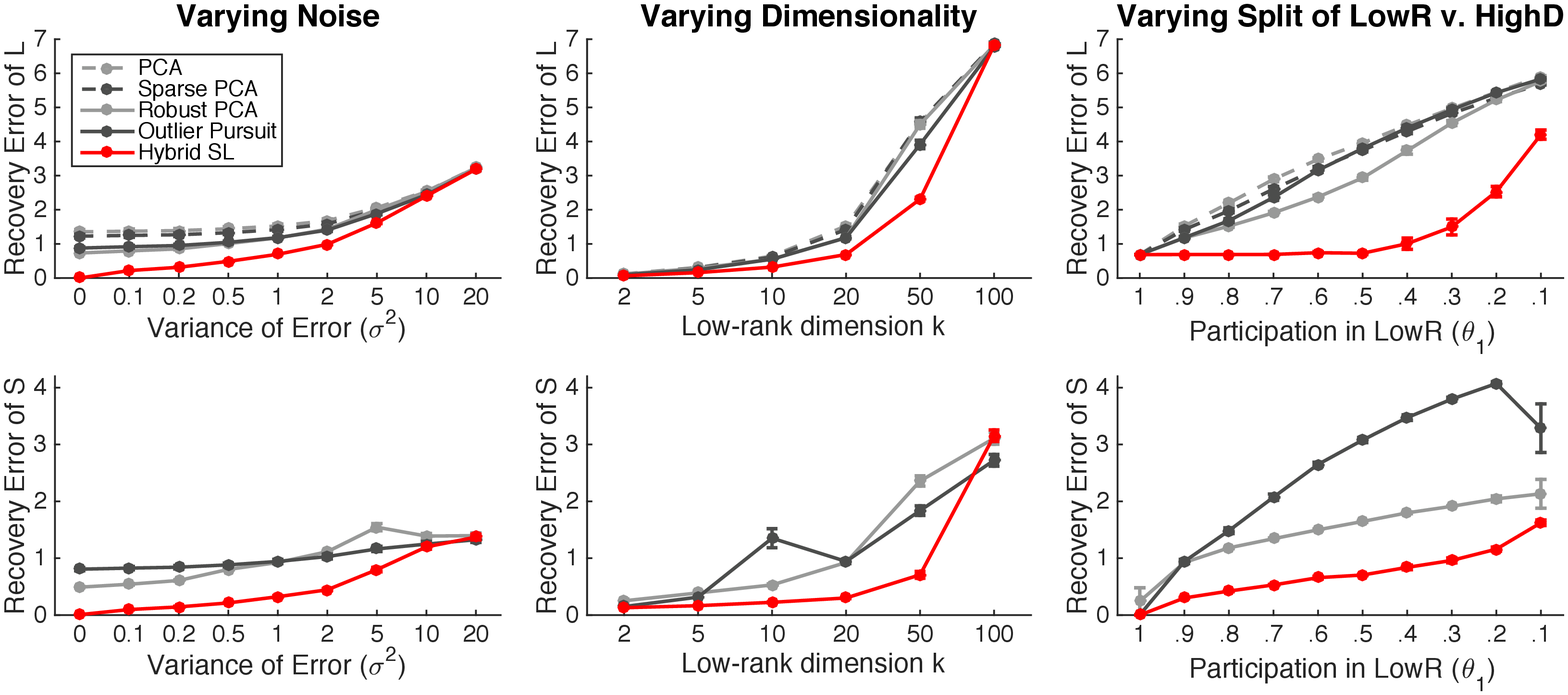}
\caption{Results comparing the performance of our hybrid model against four baselines on synthetic data.}
\label{fig:synth-exp}
\end{figure*}

In order to quantitatively evaluate our approach, we begin by performing a series of experiments on synthetic data. We compare HSL against four baseline methods that perform different variants of subspace learning: PCA, Sparse PCA~\cite{zou2006sparse}, Robust PCA~\cite{candes2011robust}, and Outlier Pursuit~\cite{xu2010robust}. Note that we apply Outlier Pursuit to the transposed data matrix, $\Xb^T$.%PCA~\cite{jolliffe2002principal}, Sparse PCA~\cite{zou2006sparse}, Robust PCA~\cite{candes2011robust}, and Outlier Pursuit~\cite{xu2010robust}. %Although many other variations of subspace learning models exist, most of them follow the same design principles as one of these baselines.
%We use the (cite paper for code) to implement the factorized method. Other code to cite?}

%In order to quantitatively evaluate the performance of our unsupervised method, we perform a series of experiments on synthetic data. We first evaluate our approach on synthetic data. 
%\vspace{-3mm}
\subsection{Data Generation} 
%\vspace{-2mm}

Given raw feature space dimensionality $p$, latent space dimensionality $k$, and sample size $n$, we first generate low-rank features $\Zb \sim \mathcal{N}(0,\Ib_{k \times k})$ and high-dimensional features $\Wb \sim \mathcal{N}(0, \Ib_{p \times p})$. %We then project the columns of each of these components into the $l_2$ unit ball. 
We then generate coefficients for the low-r component $\Ab$ by drawing uniform random values in $[-1.5,-0.5] \cup [0.5,1.5]$ and similarly generate coefficients for the high-d component $\bb$ by drawing uniformly at random from $\sqrt{k} [-1.5 ,-0.5] \cup \sqrt{k}[0.5,1.5]$. 

Next, given a probability vector $\theta = (\theta_1,\theta_2,\theta_3)$ whose elements denote the likelihood that a feature will participate in only the low-r component ($\theta_1$), only the high-d component ($\theta_2$), or both ($\theta_3$), we incorporate sparsity by setting randomly chosen columns of $\Ab$ and elements of $\bb$ to zero according to the proportions specified in $\theta$. Finally we generate the data according to $\Xb = \Zb \Ab + \Wb \diag(\bb) + \Eb$, where $\Eb \sim \mathcal{N}(0, \sigma^2)$ is i.i.d.~Gaussian noise.

\subsection{Experimental Results} 

\begin{figure*}[tbp]
\centering 
\includegraphics[width=0.68\linewidth]{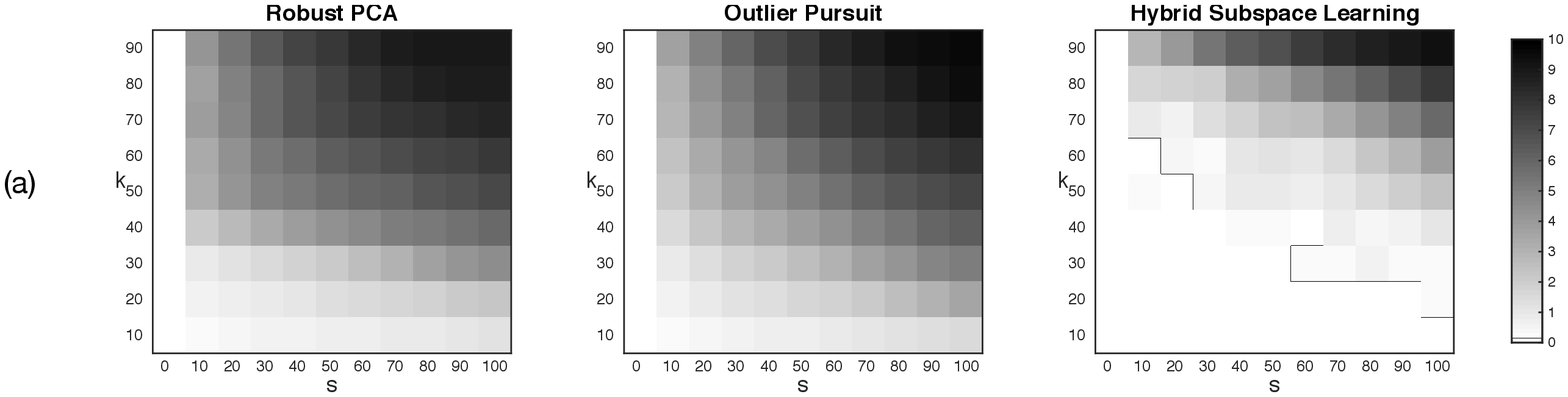} 
 \includegraphics[width=0.68\linewidth]{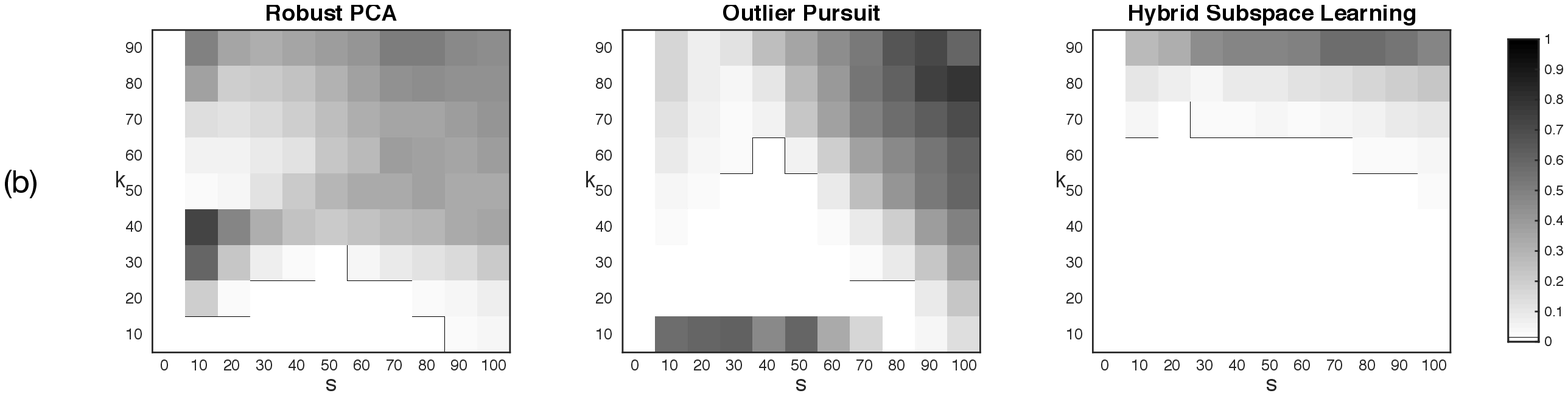} 
  \includegraphics[width=0.68\linewidth]{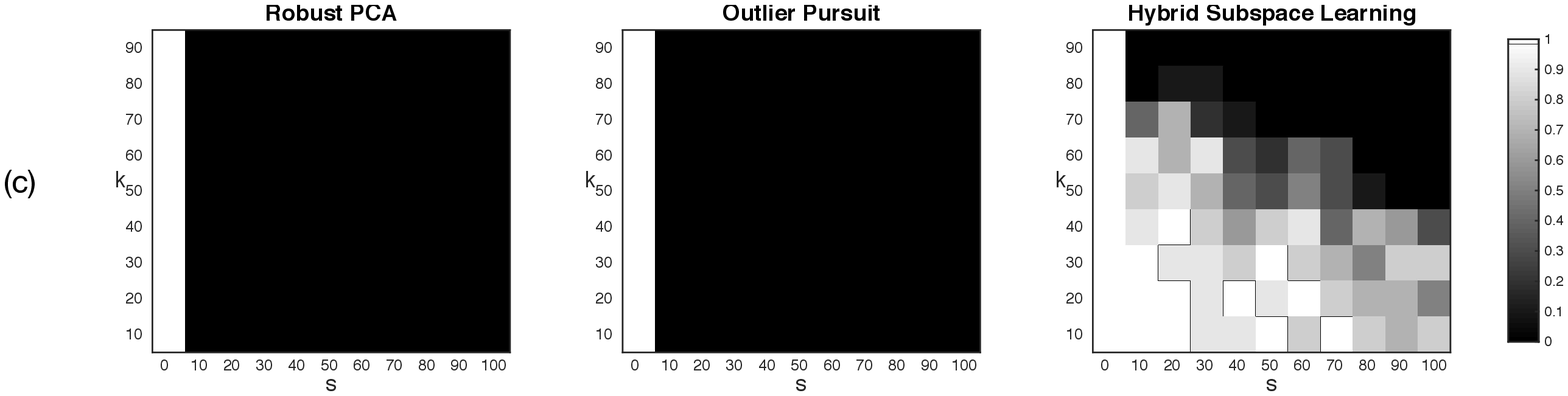}
\caption{Results of a phase transition experiment with varying $k$ and $s$.}
\label{fig:phase-transition}
\end{figure*}

\begin{figure*}
\centering
   \includegraphics[width=0.4\linewidth]{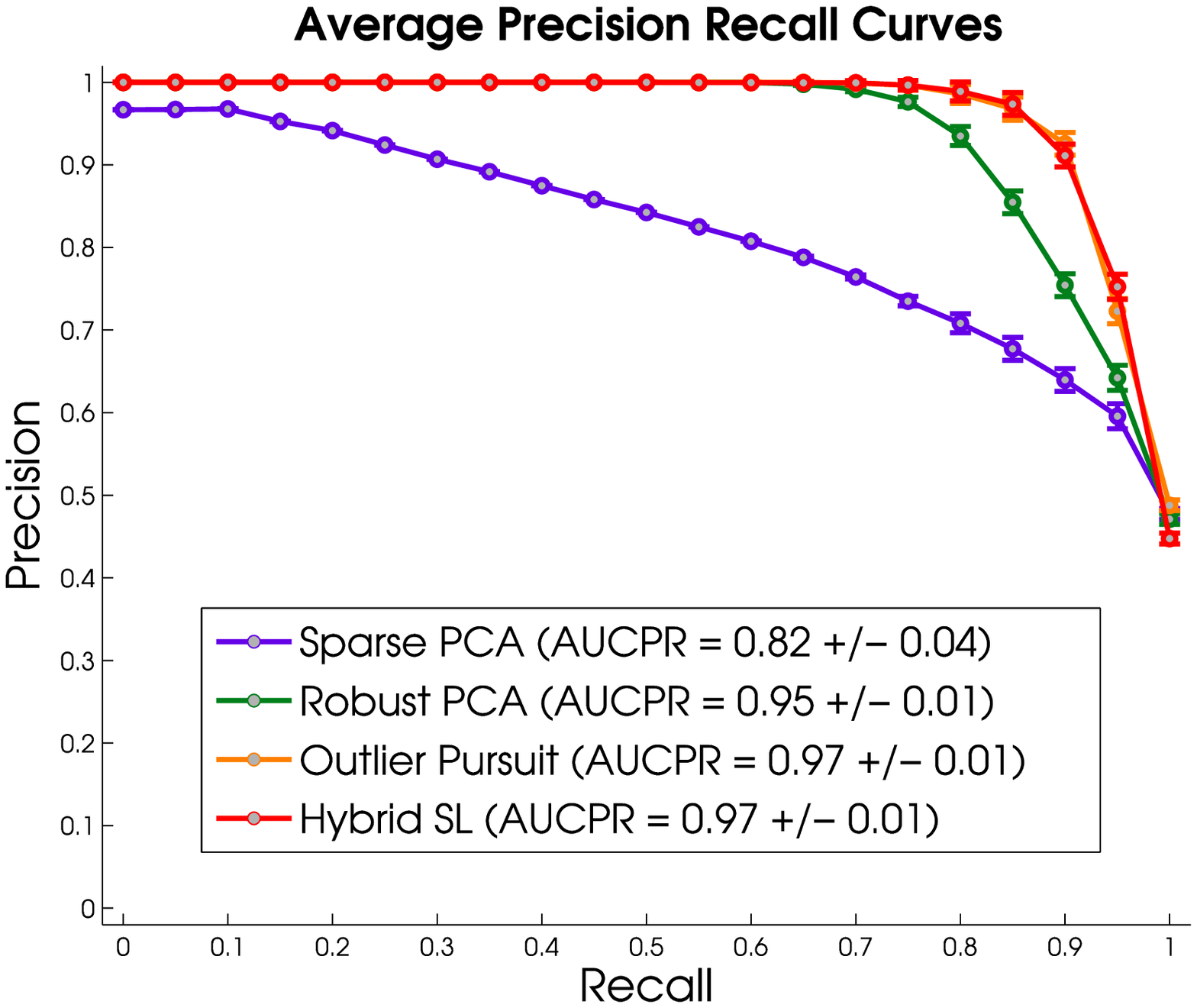}
   \includegraphics[width=0.4\linewidth]{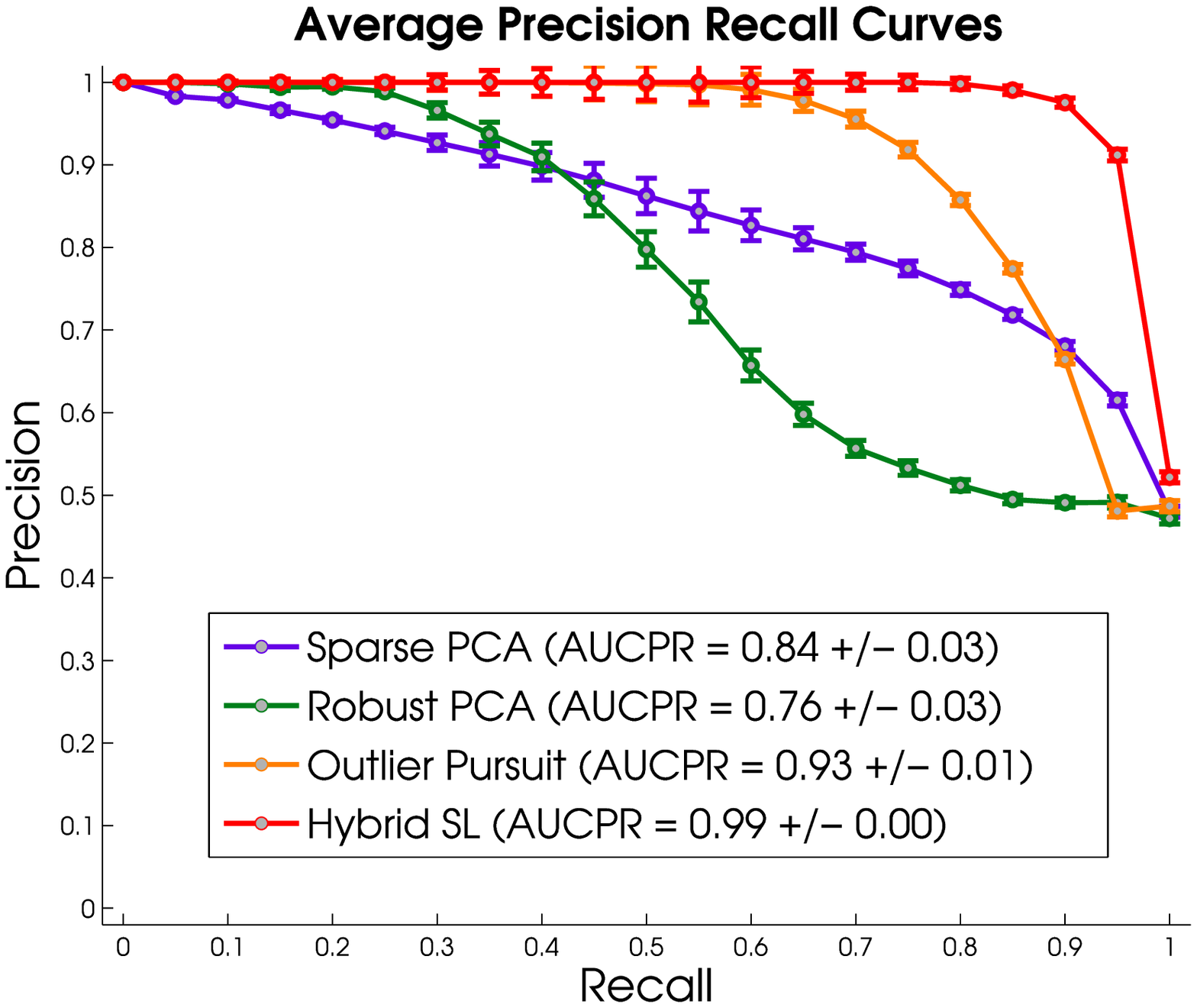}
   \caption{Precision-recall curves for SPCA, RPCA, OP, and HSL calculated by varying parameter values over a broad range and evaluating recovery of the true set of high-dimensional features.}
   \label{fig:pr-curve}
\end{figure*}

\begin{figure*}
\centering
   \includegraphics[width=0.24\linewidth]{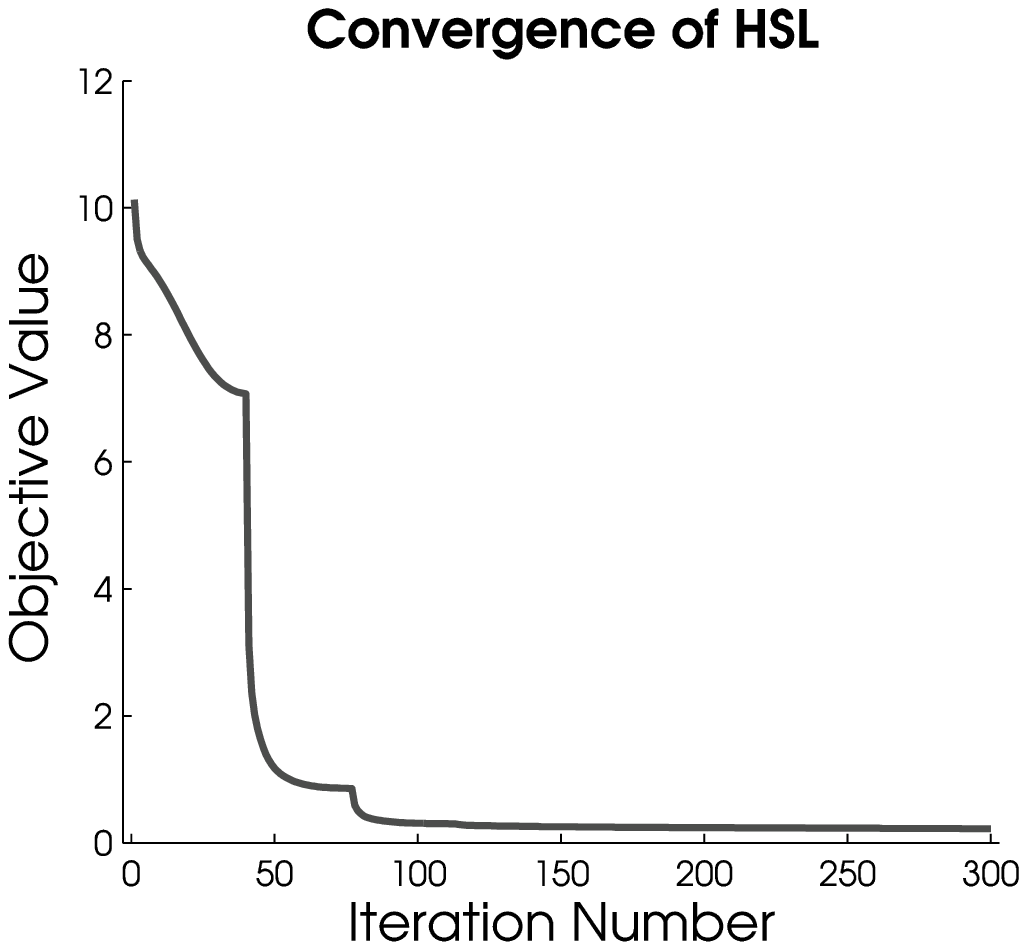}
   \includegraphics[width=0.24\linewidth]{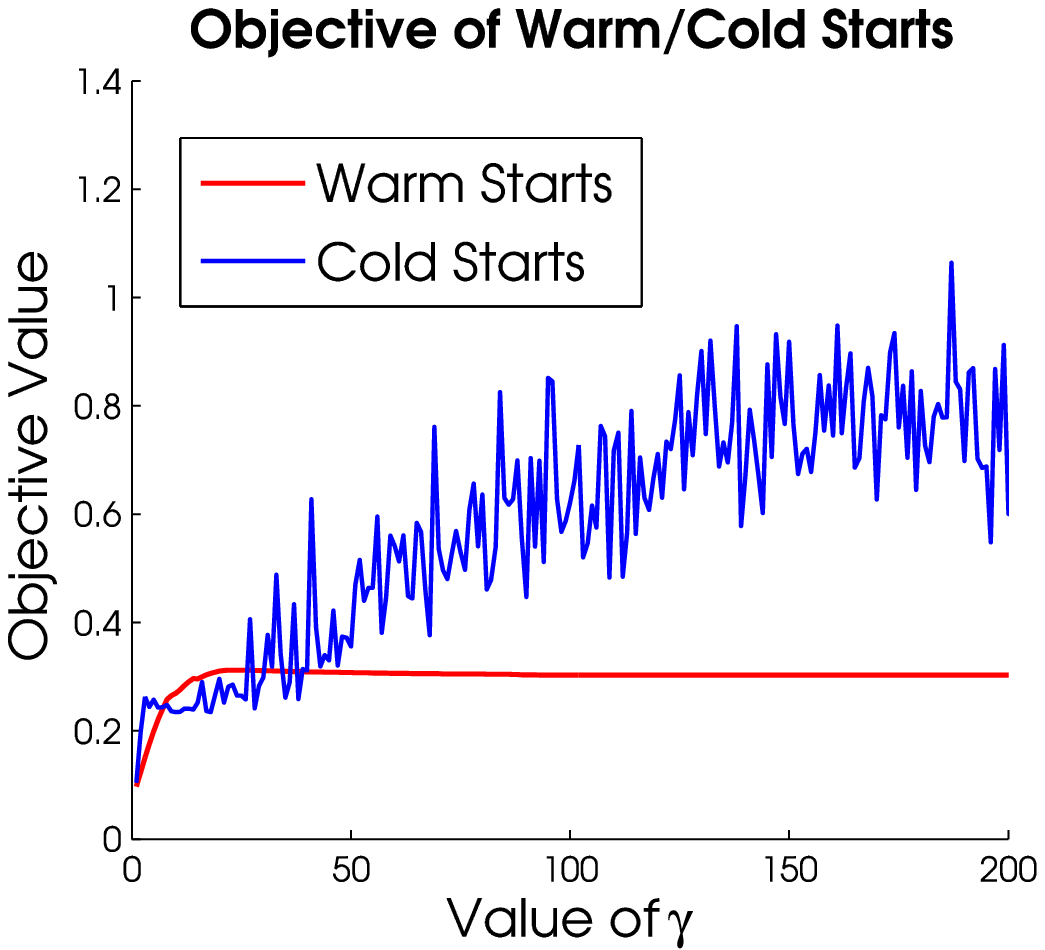}
   \includegraphics[width=0.24\linewidth]{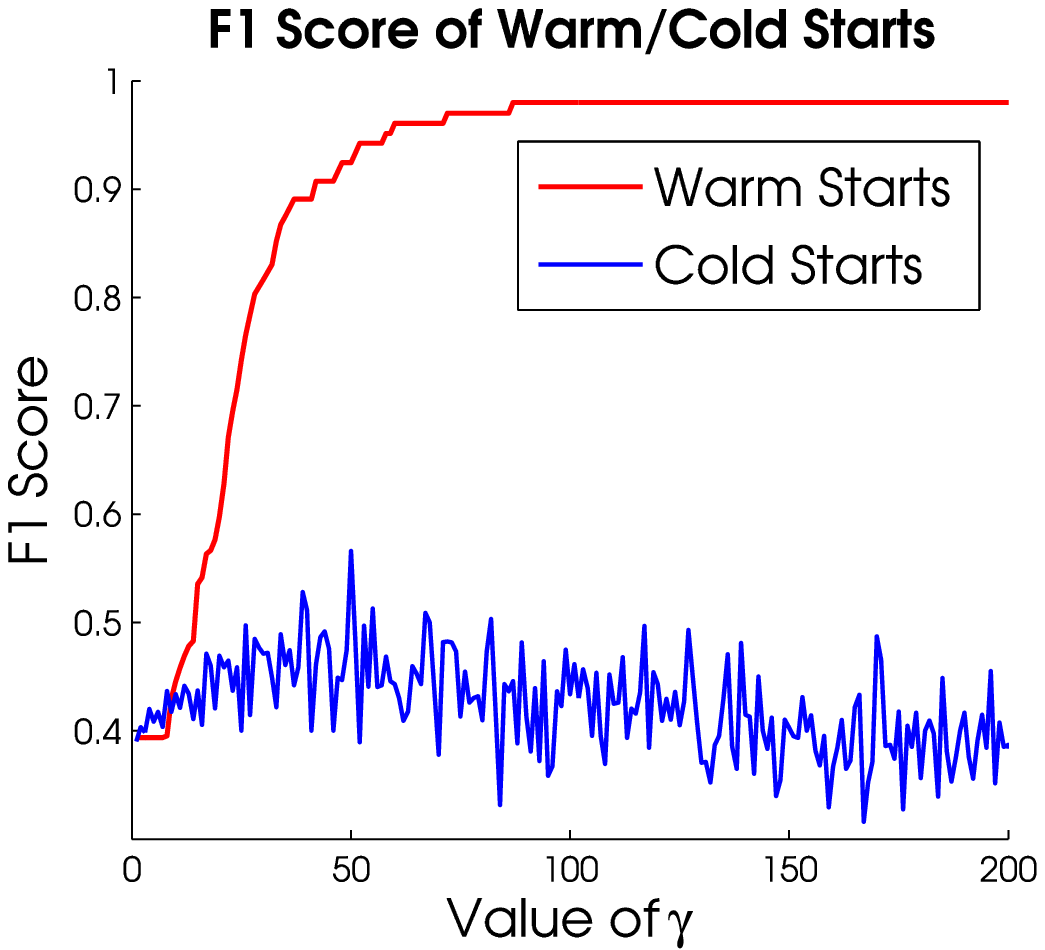}
   \includegraphics[width=0.24\linewidth]{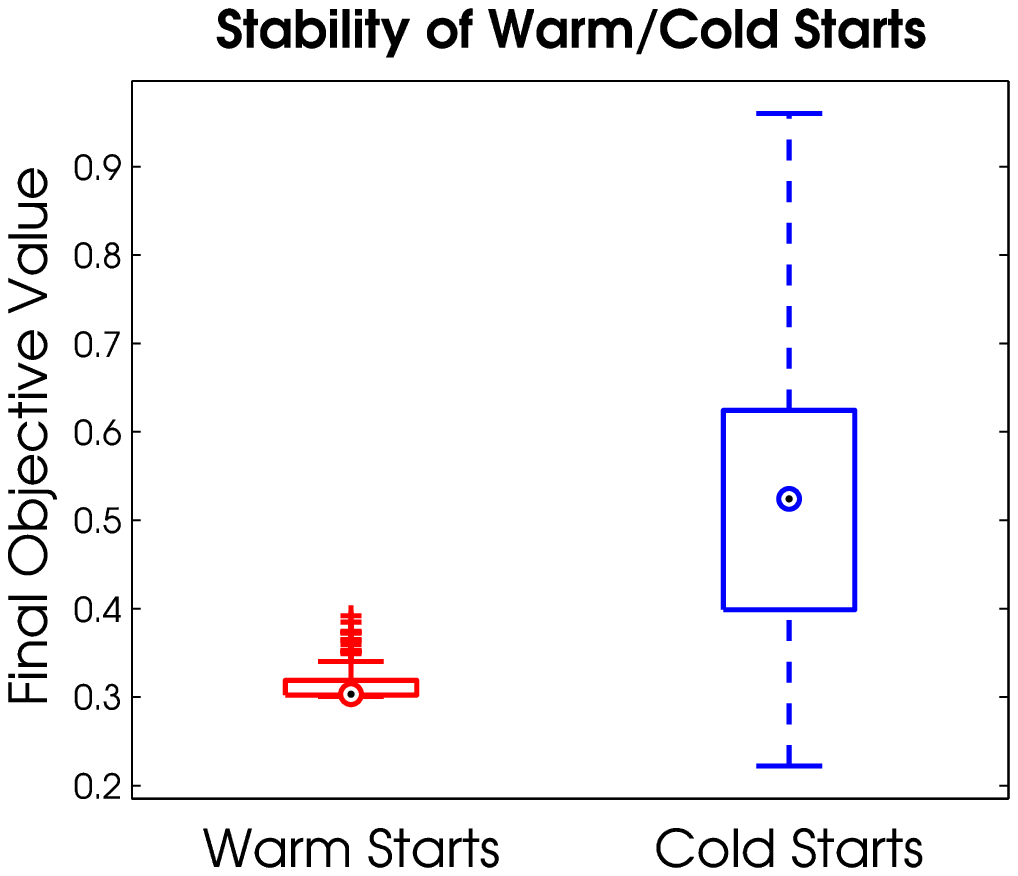}
\caption{(a) Convergence of HSL. (b) Final objective value after running HSL with each value of $\gamma$ using warm and cold starts. (c) F1 score after running HSL with each value of $\gamma$ using warm and cold starts. (d) Final objective value of HSL averaged over multiple simulations with warm and cold starts.}
\label{fig:warmstart-exp}
\end{figure*}

We compare the performance of our method against the baselines on three tasks: recovery of the true low-rank subspace, recovery of the true high-dimensional component, and identification of the set of true high-dimensional features. For each method, we evaluate recovery of the subspace based on the estimate of the low-rank matrix $\hat{\Lb}$, which is directly output by RPCA and OP, and is calculated according to $\hat{\Lb} = \hat{\Zb} \hat{\Ab}$ for SPCA and HSL. Specifically, we compare the operator $\Vb$ that projects $p$-dimensional points to the true $k$-dimensional subspace to the projection operator $\hat{\Vb}$ for the $k$-dimensional subspace closest to $\hat{\Lb}$ for each method. Similarly, we evaluate recovery of the high-dimensional component based on the estimate of the sparse matrix $\hat{\Sb}$, which is again directly estimated by RPCA and OP, and is calculated according to $\hat{\Sb} = \hat{\Wb}  \diag(\hat{\bb})$ for HSL.\footnote{PCA and SPCA do not produce a high-dimensional component, so we do not evaluate this metric for those two methods.} We measure recovery of the true $\Sb$ using the Frobenius norm distance, and we measure the recovery of the high-dimensional feature set using the F1 score. 
%We also report the total reconstruction of the data $\Xb$ for each model. %  encoded $\Lb = \Zb \Ab$ encoded by $\Sb = \Wb \diag (\bb)$
% the recovery of $\Xb$, which we measure using the total reconstruction error according to each model, the recovery of the correct low-rank latent structure, which we measure using the difference between the estimated and true value of $\Zb \Ab$, and the recovery of the correct sparse structure, which we measure using the difference between the estimated sparse component and the true value of $\Wb \bb$. 
Since parameter selection is a challenging task in unsupervised learning, each method is run with the ground truth value of $k$, and tuning parameters are chosen by picking the values that yield the best recovery of the low-rank subspace. We believe this provides a fair comparison of all methods.  
 
In our first set of experiments, we choose default parameters $n=100$, $p=200$, $k=20$, $\sigma^2=1$, $\theta=(0.9,0.1,0)$, and then vary certain parameters across a range of values in order to evaluate the performance of our model under a broad variety of settings. In particular, we vary (a) the noise $\sigma^2$, (b) the dimensionality of the latent and feature space $k$, and (c) the proportion of low-r and high-d participation ($\theta_1$ v. $\theta_2$) with no overlap. 
%, and (d) the amount of overlap between the low-r and high-d components ($\theta_3$) while keeping $\theta_1$ and $\theta_2$ equal. 
In all cases, we run HSL with $\gamma \rightarrow \gamma_{\text{max}}$ to ensure no overlap between the low-r and high-d components. The results of these experiments are shown in Figure~\ref{fig:synth-exp}. The top row shows the recovery error for the low-rank component $\Lb$, and the bottom row shows the recovery error for the high-dimensional component $\Sb$. Each point represents the mean value over 10 random datasets, and the error bars show the standard error over these trials. The results demonstrate that HSL significantly outperforms all baselines in nearly all conditions. %Since parameter selection is a challenging task in unsupervised subspace learning, tuning parameters for all models were chosen by picking the values that yield the best recovery of the shared low-rank structure $\Zb \Ab$. We believe this provides a fair comparison of all methods.  
%We note that RPCA and OP always achieve perfect reconstruction of the training data, but this is simply due to model design and does not indicate a better estimate of the underlying structure of the data.

%The results demonstrate that our method outperforms all baselines. In particular, we notice that the performance gap on the full reconstruction error and the low-rank structure recovery widens as the signal-to-noise ratio drops and as the dimensionality of the problem increases, both of which are important settings that frequently appear in real-world datasets. Although the parameters were tuned to maximize performance on the recovery of the shared latent space, we also perform significantly better than competing methods on the reconstruction of the full dataset $\Xb$. Furthermore, we still outperform all baselines in the $\theta_3 > 0$ case, despite the fact that we are setting $\gamma_{max}$ such that there is no overlap in our final estimate.

Next we perform an experiment to evaluate the phase transition of our model in the zero noise case relative to RPCA and OP. Here we use $n = 100$, $p = 200$, $\sigma^2 = 0$, and we vary the low-rank dimensionality $k$ and the number of features in the high-dimensional component, denoted by $s$. For each parameter setting, we run 10 trials and report (a) the average recovery error on $\Lb$, (b) the average selection error on $\Sb$ ($1 -$ F1 score), and (c) the average number of successes, where we define success as exactly recovering the true subspace (error $\leq .001$) and identifying the correct set of high-dimensional features ($F1 = 1.0$). The phase transition diagrams are shown in Figure~\ref{fig:phase-transition}. In the top two panels, a higher error is denoted by a darker color. HSL achieves low error in the majority of cases, whereas RPCA and OP have significantly higher error even when $k$ and $s$ are both small. In the bottom panel, white indicates success and black indicates failure. This figure shows that only HSL succeeds on both tasks (recovery of $\Lb$ and $\Sb$) when $s > 0$.

We also show a comparison of the precision-recall curves for the recovery of the high-dimensional features obtained by varying the parameter values for SPCA, RPCA, OP, and HSL over a broad range in Figure~\ref{fig:pr-curve}. The left panel shows the PR curve generated using the standard data generation approach that we previously described. Although HSL achieves a very high AUC, several other methods perform just as well. In order to make this task more challenging, we generated a second type of dataset in which the average variance of the high-dimensional features is about half the average variance of the low-rank features, making them harder to identify. The right panel shows the PR curve generated from this data. In both cases, the PR curves are averaged over 20 simulations.

Finally, we perform an empirical analysis of the effects of using cold starts versus warm starts to optimize our model. Given a dataset and a fixed value of $\lambda$, we train our model in one of two ways. Using cold starts, we simply test a series of successive values of $\gamma$, randomly initializing the model each time, until we hit $\gamma_{\text{max}}$. Using warm starts, we start with $\gamma = 0$ and increase its value incrementally, each time initializing the model with the estimate obtained on the previous value of $\gamma$, and again stop when we reach $\gamma_{\text{max}}$. As previously stated, $\gamma_{\text{max}}$ is not fixed a priori, but is chosen to be the smallest value that yields zero overlap between the low-r and high-d components. We compare the final objective value and F1 score obtained after optimizing our objective with each value of $\gamma$. The results are shown in Figure~\ref{fig:warmstart-exp}, and illustrate that using warm starts helps avoid local optima and leads to increased stability. Figure~\ref{fig:warmstart-exp} also shows that HSL with warm starts exhibits good convergence properties.

\section{Real Data Experiments}

In this section, we apply hybrid subspace learning in two different domains in order to showcase its capabilities and performance on real-world datasets.

\subsection{Background Subtraction in Videos}

\begin{figure*}[tbp]
\centering
   \includegraphics[width=0.85\linewidth]{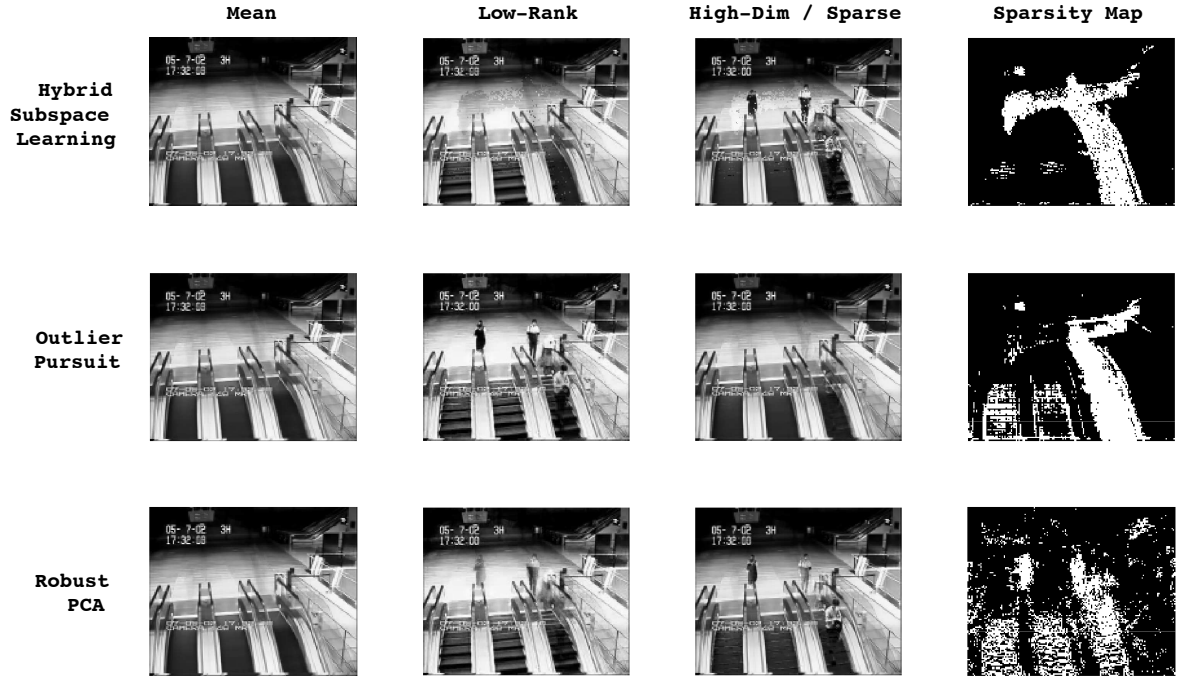} \hspace{0.5cm}
\caption{Results on one frame of the escalator dataset.}
\label{fig:real-escalator}
\end{figure*}

As an illustrative example, we begin by applying HSL to the problem of background subtraction in videos. In contrast to most traditional methods for background subtraction, which aim to distinguish between the foreground and background pixels in each frame, our results demonstrate that HSL is useful for identifying locations of consistent but irregular movement in videos. %For example, [discuss the particular video we apply to and say that pedestrians walking is irregular movement, but the escalators correspond to regular movement]

%To illustrate this concept, 
In this experiment, we applied HSL along with RPCA and OP to a video of three escalators in a subway station. In the clip, one of the escalators has considerable traffic while the other two are largely unused. However, all three escalators are running. Using $k=5$, HSL assigns nearly all of the background pixels to the low-r component, including those for the two moving but empty escalators, and assigns the remaining pixels, most of which correspond to locations with foreground movement, to the high-dimensional component. In this case, the two empty escalators exhibit {regular} movement, whereas the third escalator exhibits {irregular} movement from the pedestrian activity.

The results on a single frame of the video are shown in Figure~\ref{fig:real-escalator} and compared with the results of applying RPCA and OP to the same data. Note that hyperparameters for each method were chosen to yield the same fraction of features assigned to the low-r versus high-d components (80\%~vs.~20\%, respectively). The results of all three methods on the full video sequence are also available.\footnote{See \url{https://youtu.be/Ke0AZUn4TdM}.} In this example, only HSL is able to assign all pixels containing information about the moving people in the video to the high-dimensional component. Furthermore, the sparse map produced by HSL corresponds much more closely to the regions with consistent foreground movement in the videos, namely the locations that people move through.

%\begin{table*}
%\small
%\centering
%\caption{Reconstruction errors Gene Expression, 17814 raw features (k=8).}
%\begin{tabular}{|l|c|c|c|c|}
%\hline
%Tumor Type & PCA & Robust PCA & Outlier Pursuit & HSL \\
%\hline
%Breast & $477.40$ & $212.32$ & $1213.49$ & \\
%Colon & $535.79$ & $172.21$ & $1082.86$ & \\
%GBM & \\
%Kidney & & & & &\\
%Lung & $293.75$ & $184.93$ & $1011.72$ & $\bf{152.54}$\\
%\hline
%\end{tabular}
%\label{table:reconstruction_errors_gene}
%\end{table*}

%\begin{table*}
%\small
%\centering
%\caption{Reconstruction errors Methylation, 23094 raw features, (k=6).}
%\begin{tabular}{|l|c|c|c|c|}
%\hline
%Tumor Type & PCA & Robust PCA & Outlier Pursuit & HSL \\
%\hline
%Breast & $551.54$ & $345.46$ & $1370.85$ & \\
%Colon & $358.75$ & $488.74$ & $1274.08$ &\\
%GBM & $147.12$ & $146.49$ & $473.77$ & $\bf{91.45}$\\
%Kidney & $592.50$ & $243.85$ & $1563.99$ & \\
%Lung & $276.64$ & $595.68$ & $1400.46$ &\\
%\hline
%\end{tabular}
%\label{table:reconstruction_errors_methy}
%\end{table*}

% Argument: HSL captures low-r structure (as measured by comparable reconstruction error with just low-r structure, and more distinct clustering), and features assigned to high-d component are biologically meaningful (better survival analysis and differential gene selection)

\subsection{Genomic Analysis of Cancer}
\label{sec:real-exp}

\begin{figure*}[tbp]
 \centering
 \subfigure[Raw Data, p-value=0.18]{
     \includegraphics[width=0.3\linewidth]{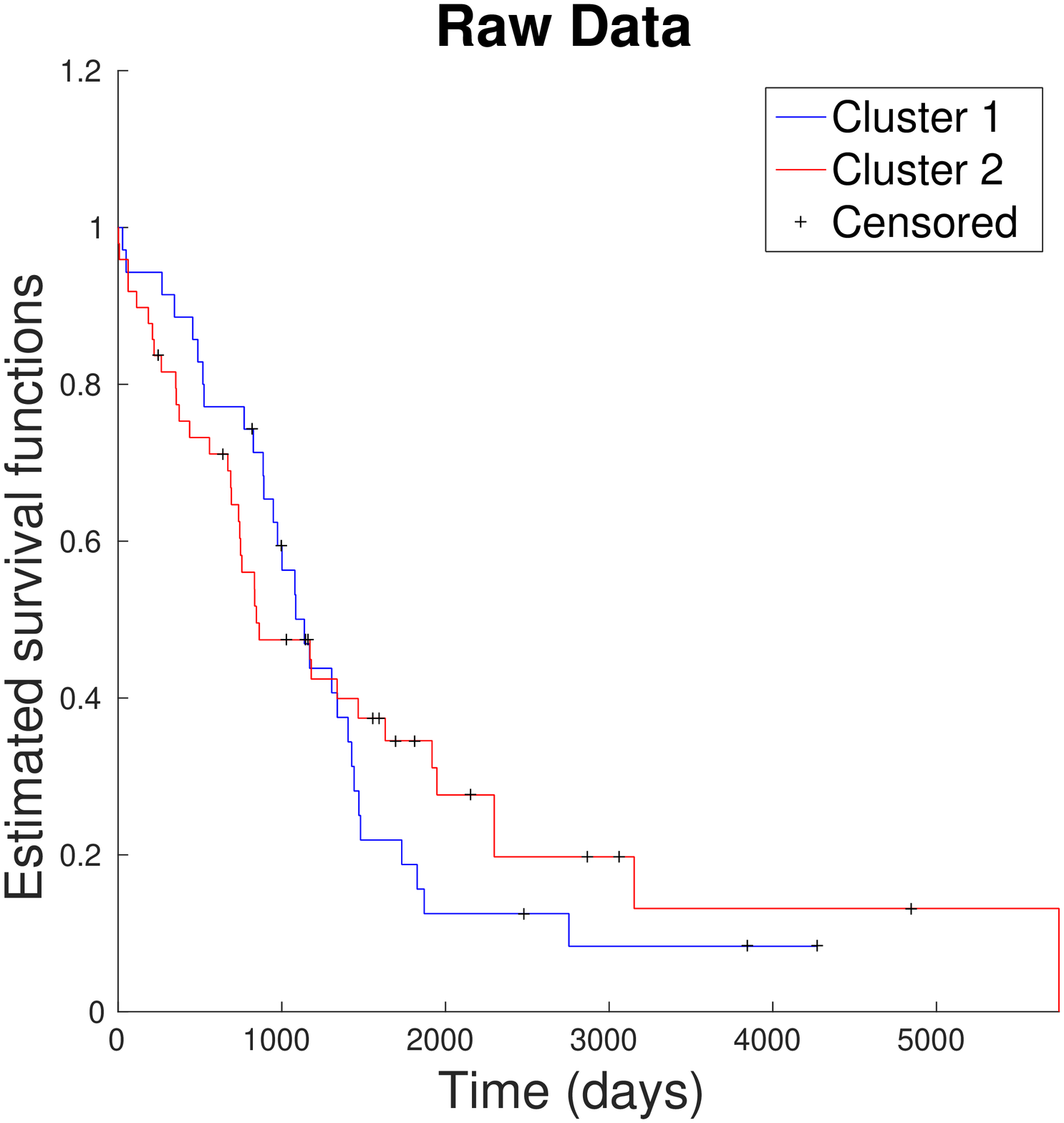}
  }  
   \subfigure[PCA, p-value=0.16]{
     \includegraphics[width=0.3\linewidth]{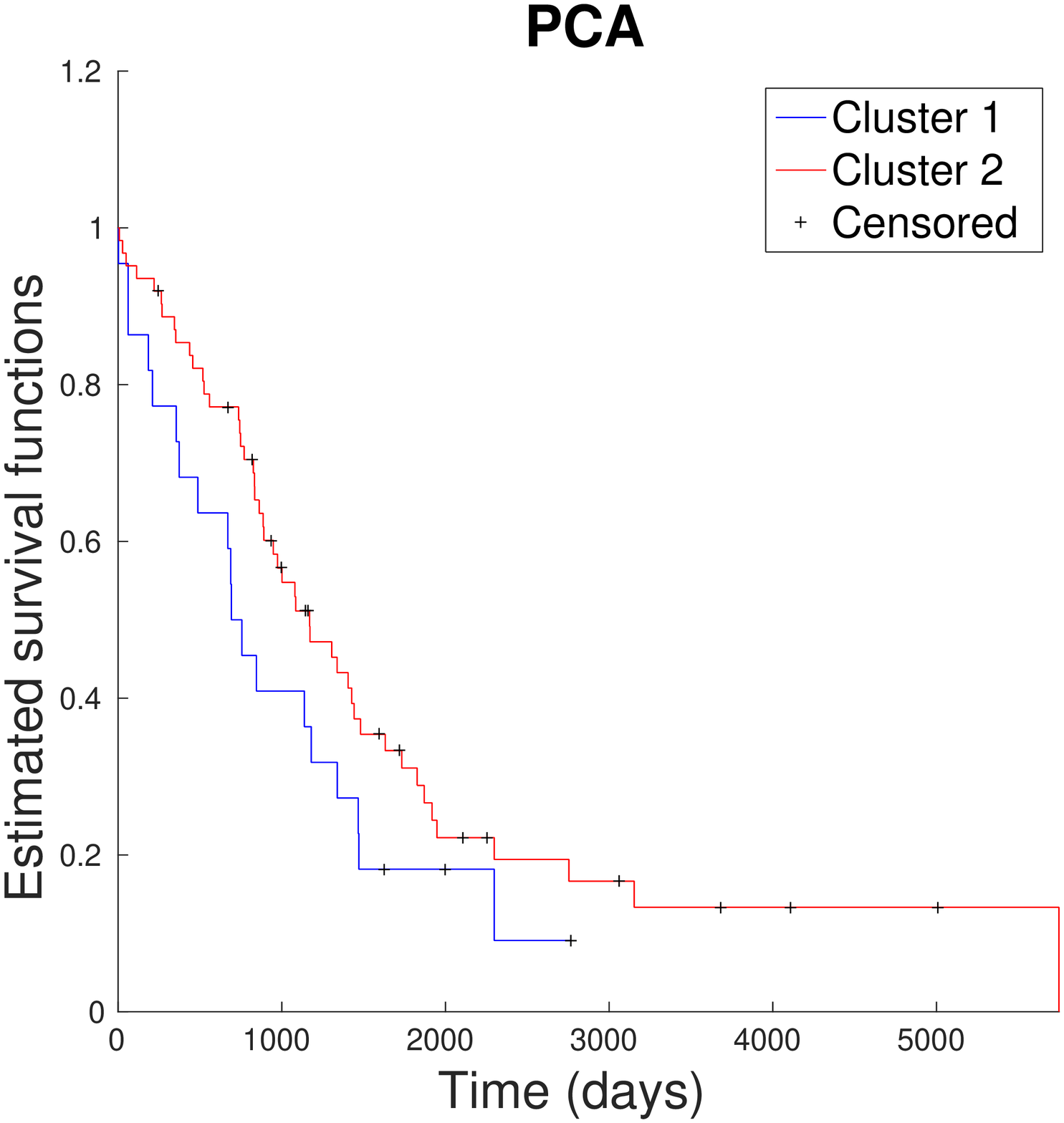}
  }  
   \subfigure[Robust PCA, p-value=0.15]{
     \includegraphics[width=0.3\linewidth]{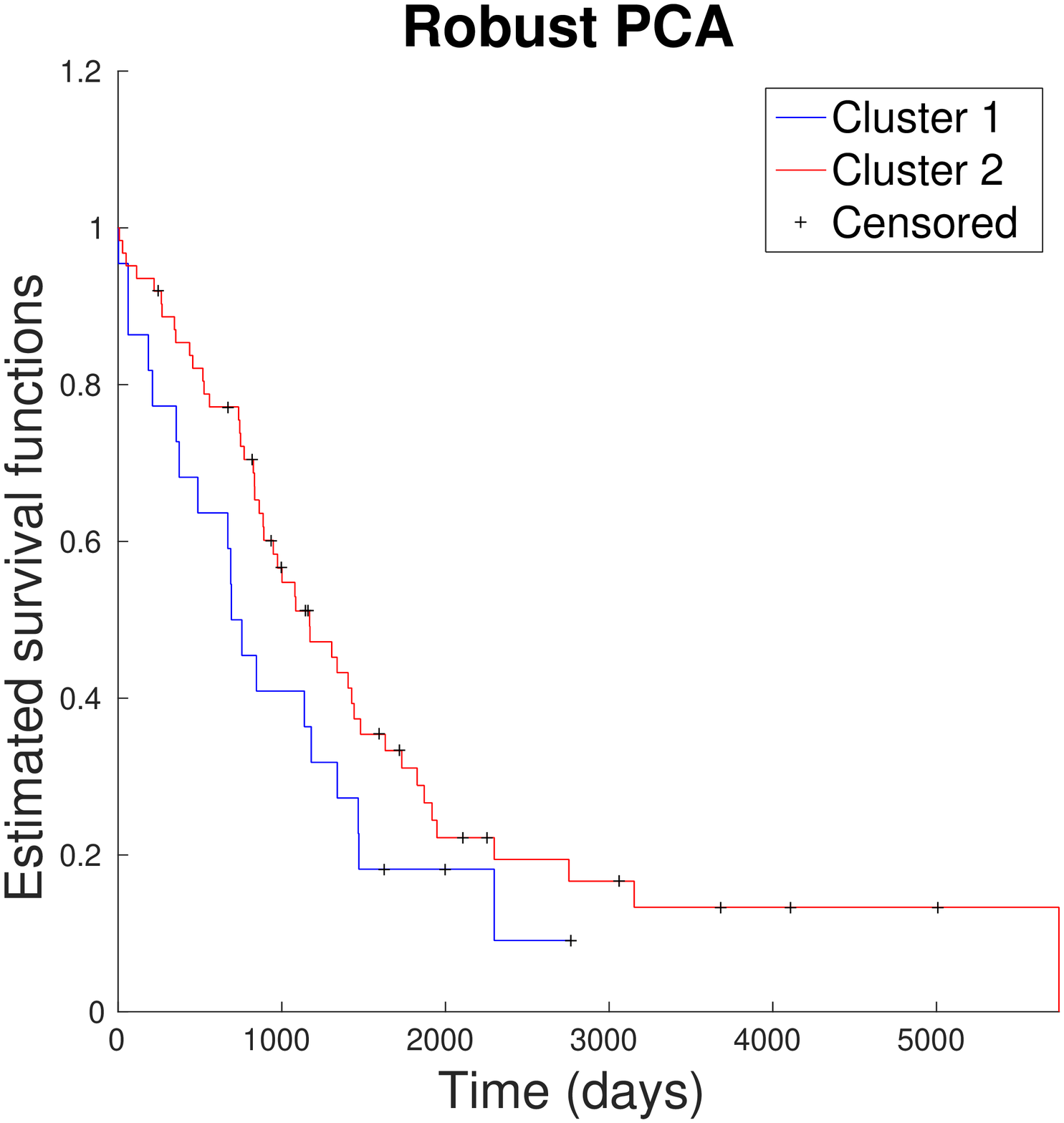}
  }  
   \subfigure[Outlier Pursuit, p-value=0.16]{
   \includegraphics[width=0.3\linewidth]{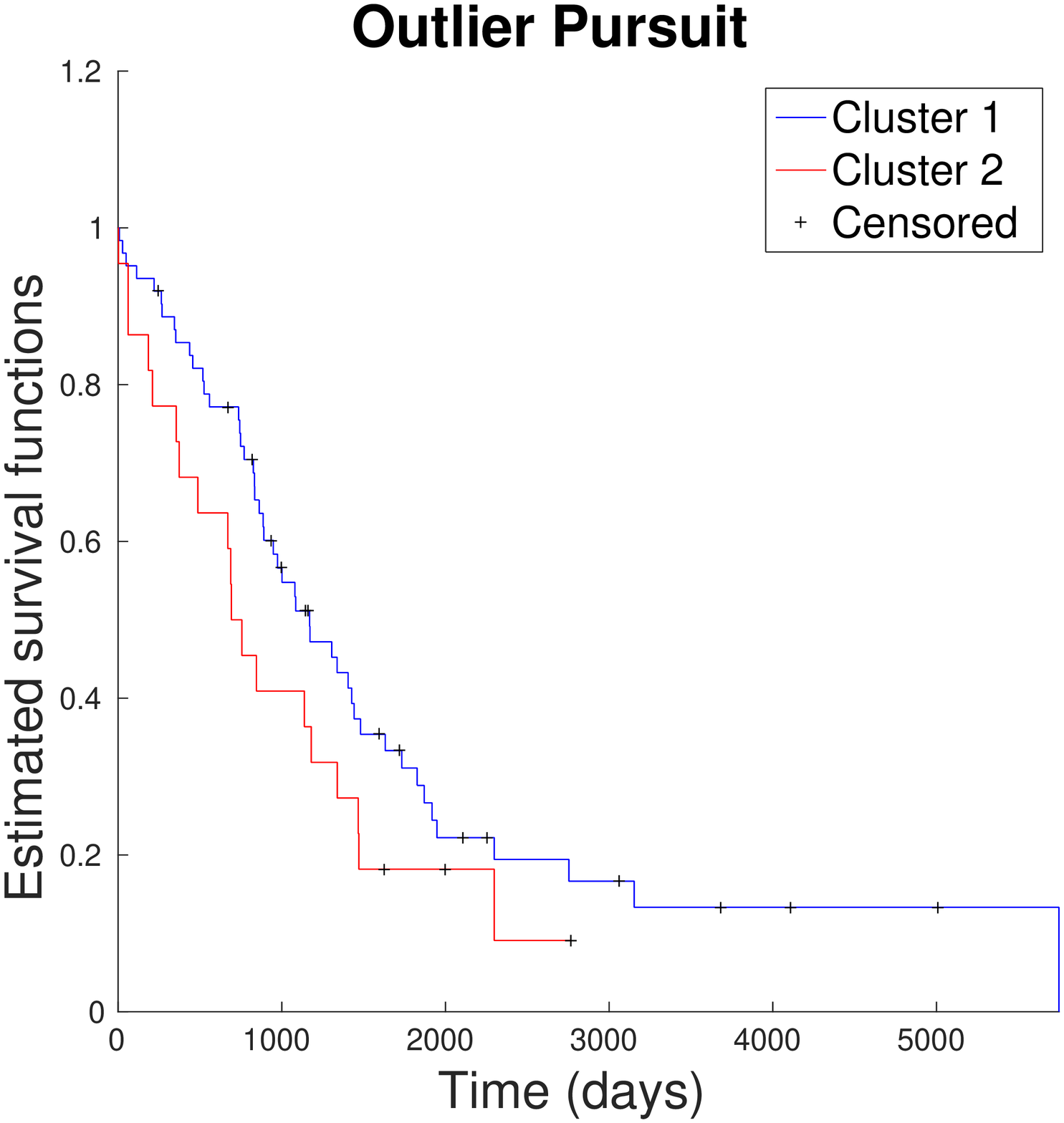}
  }  
   \subfigure[HSL, p-value=0.04]{
   \includegraphics[width=0.3\linewidth]{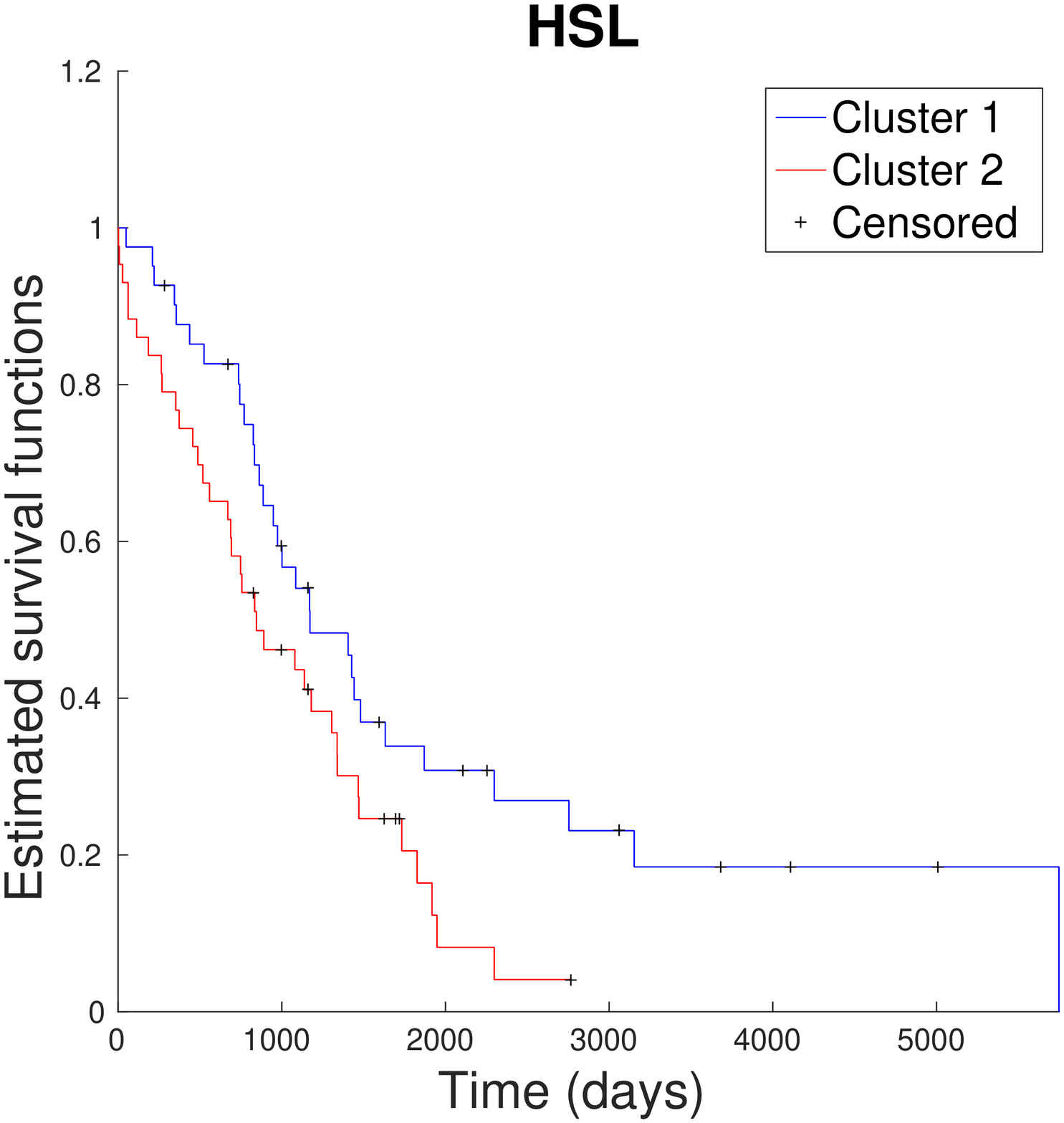}
  }
  \subfigure[Mean (std.) p-values of differential survival functions over 100 initializations of the clustering method.]{
   \includegraphics[width=0.3\linewidth]{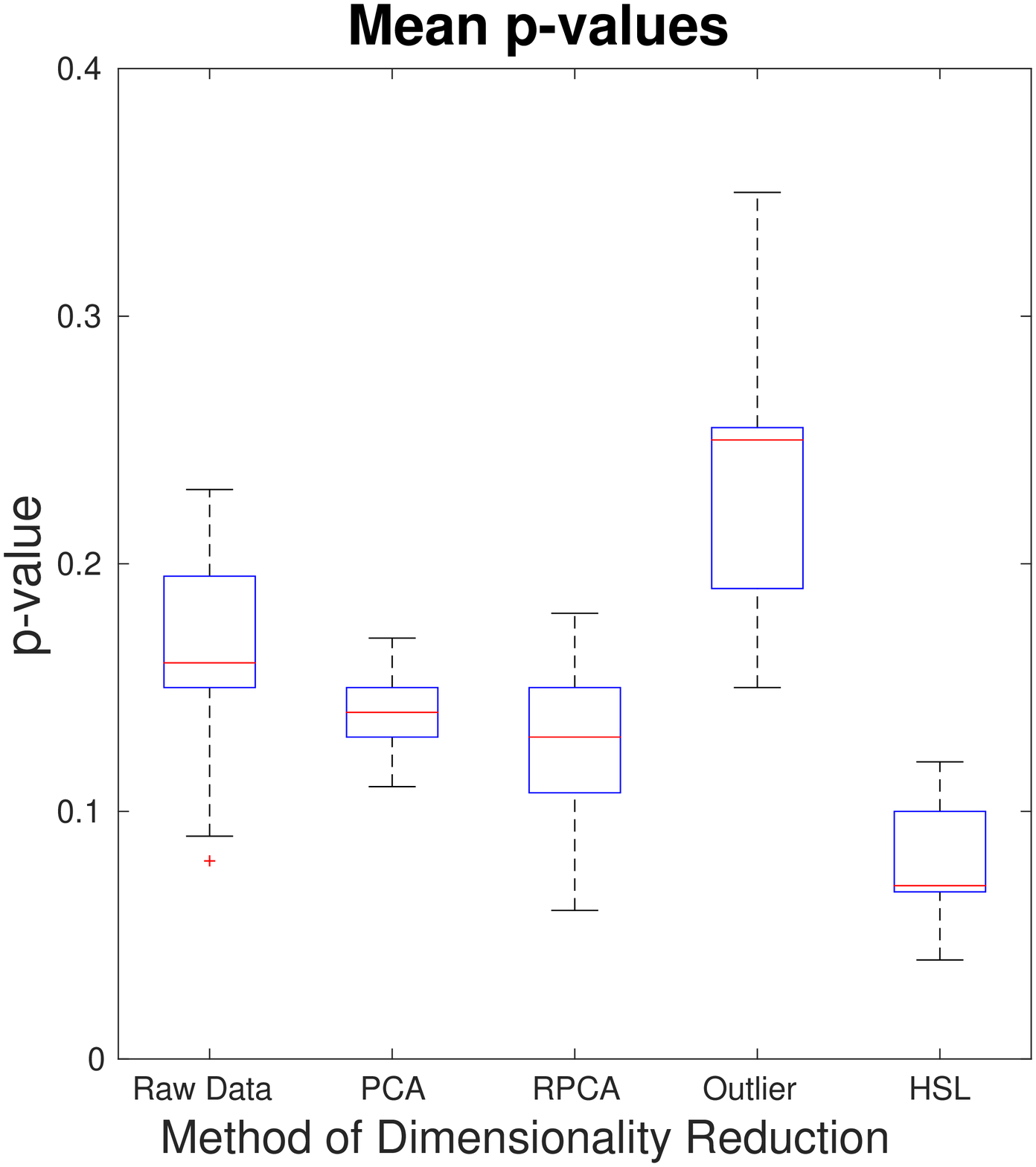}
  }  
 \caption[Survival analysis]{Survival analysis of breast cancer gene expression. (a-e) Representative Kaplan-Meier survival function estimates and corresponding p-values. (f) Distribution of p-values of differential survival functions over 100 clustering initializations.}
  \label{fig:log_cox_rank}
\end{figure*}

\begin{table*}[tbp]
\renewcommand{\arraystretch}{1.3}
\centering
\caption{Reconstruction Errors of the Low-Rank Component of miRNA Data}% after compression to 5 latent dimensions. Distance metric is the Euclidean distance.}
\vspace{0.1cm}
\begin{tabular}{|l||c||c||c||c|}
\hline
Tumor Type & PCA & Robust PCA & Outlier Pursuit & HSL \\
\hline
Breast & $63.99$ & $\bf{29.46}$ & $172.44$ & $29.61$\\ 
Colon & $83.73$ & $33.09$ & $141.17$ & $\bf{31.32}$\\
GBM & $70.35$ & $106.08$ & $303.06$ & $\bf{40.69}$\\
Kidney & $54.77$ & $45.56$ & $179.56$ & $\bf{25.93}$\\
Lung & $54.74$ & $\bf{25.31}$ & $172.97$ & $25.73$\\
\hline
\end{tabular}
\label{table:reconstruction_errors_mirna}
\end{table*}

\begin{table*}[tbp]
\renewcommand{\arraystretch}{1.3}
\centering
\caption{Silhouette Scores for Clusters Produced by $k$-Means}
\vspace{0.1cm}
\begin{tabular}{|l||c||c||c||c||c|}
\hline
Tumor Type & Raw Data & PCA & Robust PCA & Outlier Pursuit & HSL \\
\hline
Breast & $0.35 \pm 0.07$ & $0.51 \pm 0.04$& $.27 \pm .02$ & $0.17 \pm 0.02$ & $\bf{0.65} \pm 0.08$\\
Colon & $0.37 \pm 0.17$ & $0.52 \pm 0.07$ & $0.30 \pm 0.04$ & $0.15 \pm 0.05$ & $\bf{0.7}0 \pm 0.07$\\
GBM & $0.22 \pm 0.05$ & $0.45 \pm 0.06$ & $0.20 \pm 0.03$ & $0.15 \pm 0.07$ & $\bf{0.48} \pm 0.06$\\
Kidney & $0.26 \pm 0.04$ & $0.43 \pm 0.04$ & $0.24 \pm 0.02$ & $0.13 \pm 0.04$ & $\bf{0.59} \pm 0.08$\\
Lung & $0.29 \pm 0.05$ & $\bf{0.53} \pm 0.05$ & $0.28 \pm 0.03$ & $0.19 \pm 0.05$ & $0.52 \pm 0.09$\\
\hline
\end{tabular}
\label{table:silhouettes}
\end{table*}

\begin{table*}[tbp]
\renewcommand{\arraystretch}{1.3}
\caption{Differential Enrichment of the Features Assigned to High-Dimensional Components}
\vspace{0.1cm}
\begin{adjustbox}{center}
\begin{tabular}{ |c||c||c||c|} 
 \hline
Data Type & Gene Ontology Code & Gene Ontology Term & Selected Oncogenes\\
\hline 
 & GO:0032633 & Interleukin-4 production & LEF1, CD83\\ 
Tumor & GO:0017111 & Nucleoside-Triphosphatase Activity & TCIRG1, RAB31, ATP6V1C1, ATP6V1G3\\
& GO:0005515 & Protein Bnding & NTRK3, HSPA1A, CCR5, ITGA2 + 10 more\\
 \hline 
 & GO:0034472 & snRNA 3'-end Processing &  None \\
Control & GO:0005006 & Epidermal Growth Factor Receptor Activity & ERRFI1, PSEN1 \\
 & GO:0060478 & Acrosomal vesical exocytosis & None \\ 
  \hline
\end{tabular}
\end{adjustbox}
\label{table:goterms}
\end{table*}

%1. Overview of experiment and results	
Next we apply HSL to biomedical data, and provide both qualitative and quantitative results to illustrate its performance. A common biological case in which $p \gg n$ is that of microarray data, in which the number of features measured typically far exceeds the number of patients for whom data is available. Here, we study the effectiveness of applying subspace learning methods to microarray data taken from cancer patients. We show that our approach outperforms several baselines on this data. Specifically, HSL produces subspace embeddings that achieve lower reconstruction error and lead to better performance on downstream tasks than competing methods. Finally, we demonstrate that HSL can also be used as a feature selection algorithm, since the features assigned to the high-dimensional component reflect biological characteristics of the original data.

%2. Description of datasets we have
To conduct our experiments, we used two datasets from TCGA\footnote{The Cancer Genome Atlas, \url{http://cancergenome.nih.gov/.}}. The first dataset contains miRNA expression levels for five types of cancer -- breast, glioblastoma multiforme (GBM), colon, kidney, and lung. Within each cancer type, we have data for 106, 93, 216, 123, and 107 samples, respectively, and each sample has 354 miRNA features. We use this dataset to evaluate how well the low-rank embedding of HSL captures the original data and its characteristics. The second dataset contains gene expression data for breast cancer patients and matching control samples. It contains 13794 mRNA features for 106 samples. We used this to analyze the high-dimensional component of HSL and to determine whether the information contained in the HSL estimate could sufficiently differentiate between cancer and control samples. EDIT HERE
%The diversity of these data allow us to evaluate the robustness of HSL by measuring reconstruction error and clustering cohesiveness. 
%To examine the biological utility of the estimated subspaces, we required control samples and interpretable gene-level features. The only dataset in TCGA meeting these specifications was the breast cancer gene expression dataset, which contains 13794 mRNA features for 106 samples.

%3. Description of how we run on these datasets, including hyperparameter selection and other details
For each dataset, the number of latent dimensions $k$ was chosen by manually inspecting the singular value spectrum. This value was determined to be $k=5$ for the miRNA datasets and $k=30$ for the gene expression dataset. In all experiments, we selected hyperparameter values as follows. For RPCA, the value of $\lambda$ was set to $\frac{1}{\sqrt{n}}$, which can optimally recover the low-rank structure under standard assumptions \cite{xu2010robust}. In keeping with our synthetic experiments, OP was run on the transposed data matrix. The value of $\lambda$ for OP was chosen to produce a low-rank component with rank equal to $k$. For HSL, parameters were selected by performing a grid search and selecting the combination of parameters that minimized the AIC score.

%4. Results of reconstruction error experiment
In our first experiment, we evaluated the quality of the low-r components estimated for each miRNA dataset. To do this, we measured the reconstruction errors of the low-r embeddings produced by each method.
%In our first experiment, to evaluate the quality of the low-r component of the learned subspace, we measured the reconstruction error of the low-r embedding of each miRNA dataset. %From 354 miRNA expression features, we estimated $k = 5$ latent dimensions.
Reconstruction errors, calculated as the Euclidean distance between the original data $\Xb$ and the estimated low-r component $\hat{\Lb}$, are shown in Table \ref{table:reconstruction_errors_mirna}. We see that HSL performs at least comparably, and frequently outperforms, all baseline methods on all datasets.

%5. Results of clustering experiment
Next, we hypothesized that the low-r component of the HSL embedding may be more biologically informative than those estimated by traditional subspace learning methods. To study this, we used the estimated low-rank embeddings from each method to cluster the samples within each cancer type into subtypes. Since we do not have ground truth information about the subtypes, 
%(as clinical subtypes may not accurately reflect the molecular characteristics), 
we evaluated the quality of the clusters by their silhouette scores, which provide a measure of how well the samples fit into their respective clusters. We performed $k$-means clustering using $4$ clusters for breast\cite{voduc2010brcasubtypes}, GBM\cite{verhaak2010gbmsubtypes}, and colon\cite{Guinney2015Colon} cancers and $5$ clusters for kidney\cite{Prasad2006Kidney} and lung\cite{West2012Lung} cancers, where the number of clusters is based on the number of experimentally identified subtypes.
The mean and standard deviation of the silhouette scores over 100 initializations of the clustering algorithm are shown in Table \ref{table:silhouettes}. From these results, we see that the features extracted from the low-r component of the hybrid model yield more coherent clusters than features extracted from baseline methods.% As the hybrid model does not include all features in the low-rank subspace, these results indicate that samples can be effectively differentiated with only a select number of the original features.

Since our hybrid model does not encode all the features of the original data in the low-rank subspace, using these features alone would not necessarily be expected to boost performance on downstream tasks. 
%Our results indicate that a number of miRNA features do not lie neatly on a low-rank subspace. 
Furthermore, the features assigned to the high-d component of the model likely correspond to genes that display uncommon activity patterns, which is why they cannot be easily represented by the same low-rank structure as the other genes. Based on this reasoning, we hypothesized that, rather than being unimportant, some of these genes may actually have very important biological functions. This is particularly likely in the case of cancer data, since genes that are mutated in cancerous cells display highly aberrant activity that disrupts their normal correlations with other genes. 
% Argue that these features may actually be particularly interesting because they may have aberrant expression caused by mutations; so we hypothesize that we are likely to find cancerous genes among these
%If the high-d features also have low variance, PCA-based methods will retain little of the information encoded in them. This is a problem in the biological setting; the features that have low variance are often tightly regulated due to their large biological influence. 
%We hypothesized that some of the features assigned to the high-d component of the hybrid subspace model may be biologically meaningful.

%6. Results of analyzing high-dimensional component
To test this hypothesis, we investigated whether genes assigned to the high-d component in HSL are enriched for oncogenes when the model is run on cancerous samples but not enriched for oncogenes when it is run on samples of healthy tissue. For this experiment, we used the breast cancer gene expression data with matching control samples. After estimating the latent subspaces, we identified gene ontology (GO) terms by performing an enrichment analysis\cite{eden2009gorilla} of the features comprising the high-d component, and identified known oncogenes\cite{generanker} in the subsets. For both cancer and control samples, the three GO terms with the lowest p-value for each dataset, and their contained oncogenes, are shown in Table \ref{table:goterms}. % 1324 features comprising the high-d component

From these results, we see that HSL identifies a significant number of oncogenes when trained on tumor samples but selects non-oncogenic genes when trained on the healthy control samples. Notably, the high-d component estimated from the breast cancer tumor dataset selected features involved in the regulation of Interleukin-4, an enzyme that is known to be key in the growth of human breast cancer tumors \cite{nagai2000breastcancer}. In contrast, the high-d component learned from a control group did not include those features, instead assigning them to the low-rank space. In addition, the high-d component for the cancerous samples is enriched for the GO term ``nucleoside-triphosphatase activity", which includes both ATPase and GTPase activity. These processes are involved in regulation of the cell metabolism, a central mechanism in tumor growth\cite{cairns2011regulation}. Once again, the hybrid model assigned these features to the low-r component for non-cancerous samples. As the two datasets share the same set of features, the differential enrichment of oncogenes in the high-d component suggests that the assignment of features to either high-d or low-r component reflects characteristics of the original data.

%7. Results of survival analysis experiment
Finally, we studied whether the subspaces estimated by HSL are more useful for downstream analysis than those of competing methods. To do this, we clustered the low-rank embeddings estimated from gene expression levels of both tumor and control samples into two groups using $k$-means. As seen in Figure \ref{fig:log_cox_rank}, clusters formed in the subspace estimated by HSL have more differential survival patterns than clusters formed in the subspaces estimated by traditional methods. %Using 100 initializations of the clustering algorithm in each subspace, the mean and stddev p-values under the Kaplan-Meier test were: Raw Data, 0.15 (0.02); PCA, 0.19 (0.03); RPCA, 0.15 (0.01); OP, 0.16 (0.01); HSL, 0.07 (0.02).
While the survival effect size is not large, HSL is the only dimensionality reduction technique that retains enough information to produce any survival curves that are different at a significance level of $p<.05$. This indicates that the subspace estimated by HSL is not only efficient, but also retains information for downstream analysis.

%\vspace{-0.1cm}
\section{Conclusion}
\label{sec:conclusion}
In this work, we present a new subspace learning model. Our approach employs a novel regularization scheme to estimate a partial low-rank latent space embedding of a high-dimensional dataset, and simultaneously identifies features that do cannot easily be embedded in a low-rank space. This model addresses a critical gap in the existing literature on subspace learning, in which it is usually assumed that the high-dimensional data can be completely captured by a low-rank approximation, modulo some noise. 

By comparing the singular value decompositions of real and synthetic datasets, we demonstrate that this assumption is not fulfilled in many real datasets. We therefore argue that our model is more appropriate for subspace learning on high-dimensional datasets that have a long-tailed singular value spectrum. 
%This assumption does not always hold, especially in biological systems in which influential features may be constrained through tight feedback loops to have low variance.
Through applications to synthetic data, a video background subtraction task, and real gene expression data, we demonstrate that hybrid subspace learning can effectively learn a low-rank latent structure while assigning meaningful features to the high-dimensional component. 

% Can use something like this to put references on a page
% by themselves when using endfloat and the captionsoff option.
\ifCLASSOPTIONcaptionsoff
  \newpage
\fi

\bibliographystyle{IEEEtran}
\bibliography{references}

\end{document}